\documentclass[acmtog]{acmart}

\usepackage{multirow}
\usepackage[table]{xcolor}
\usepackage{xspace}
\usepackage{bm}
\usepackage[svgnames]{xcolor}      
\usepackage{cuted}    
\usepackage{enumitem}
\usepackage{booktabs}
\usepackage{siunitx}      
\usepackage{wrapfig}
\usepackage{pifont}             
\usepackage{listings}
\usepackage{array}
\usepackage{float}
\usepackage[utf8]{inputenc}
\usepackage[T1]{fontenc}

\definecolor{siggold}{RGB}{255,215,0}  
\definecolor{sigbronze}{RGB}{205,127,50}
\definecolor{sigred}{HTML}{DD2828}
\definecolor{sigblue}{HTML}{2D2DDD}
\definecolor{siggray}{HTML}{D5D5D5}
\definecolor{sigteal}{RGB}{0,110,120}

\newcommand{\softbold}[1]{{\bfseries\color{black!80}#1}}

\newcommand{\bestMT}[1]{\hspace{3pt}\cellcolor{siggold!16}{\softbold{\num{#1}}}}
\newcommand{\best}[1]{\cellcolor{siggold!16}{\softbold{\num{#1}}}}
\newcommand{\second}[1]{\cellcolor{sigbronze!14}\num{#1}}

\usepackage{comment}

\def\eg{\emph{e.g.,}\xspace}
\def\ie{\emph{i.e.,}\xspace}
\def\etal{\emph{et al.}\xspace}

\newcommand{\cmark}{\ding{51}}     
\newcommand{\xmark}{\ding{55}}   

\newcommand{\HaMerP}{HaMeR$^{\mathrm{D}}$}

\AtBeginDocument{%
  }

\copyrightyear{2026}
\acmYear{2026}
\setcopyright{cc}
\setcctype{by-nc-nd}
\acmConference[SIGGRAPH Conference Papers '26]{Special Interest Group on
Computer Graphics and Interactive Techniques Conference Conference
Papers}{July 19--23, 2026}{Los Angeles, CA, USA}
\acmBooktitle{Special Interest Group on Computer Graphics and
Interactive Techniques Conference Conference Papers (SIGGRAPH Conference
Papers '26), July 19--23, 2026, Los Angeles, CA, USA}
\acmDOI{10.1145/3799902.3811047}
\acmISBN{979-8-4007-2554-8/2026/07}

\acmSubmissionID{216} 

\begin{document}

\title{
  \hspace{-8pt} \textsc{EgoForce}: Forearm-Guided Camera-Space 3D Hand Pose from  a Monocular Egocentric Camera
}

\author{Christen Millerdurai}
\email{Christen.Millerdurai@dfki.de}
\orcid{0009-0001-1653-8126}
\affiliation{%
  \institution{Deutsches Forschungszentrum f{\"u}r K{\"u}nstliche Intelligenz (DFKI)}
  \city{Kaiserslautern}
  \country{Germany}
}

\author{Shaoxiang Wang}
\email{Shaoxiang.Wang@dfki.de}
\orcid{0009-0006-0683-4200}
\affiliation{%
  \institution{Deutsches Forschungszentrum f{\"u}r K{\"u}nstliche Intelligenz (DFKI)}
  \city{Kaiserslautern}
  \country{Germany}
}

\author{Yaxu Xie}
\email{Yaxu.Xie@dfki.de}
\orcid{0009-0008-8345-2825}
\affiliation{%
  \institution{Deutsches Forschungszentrum f{\"u}r K{\"u}nstliche Intelligenz (DFKI)}
  \city{Kaiserslautern}
  \country{Germany}
}

\author{Vladislav Golyanik}
\email{golyanik@mpi-inf.mpg.de}
\orcid{0000-0003-1630-2006}
\affiliation{%
  \institution{Max Planck Institute for Informatics (MPII)}
  \city{Saarbr{\"u}cken}
  \country{Germany}
}

\author{Didier Stricker}
\email{didier.stricker@dfki.de}
\orcid{0009-0004-8794-6858}
\affiliation{%
  \institution{Deutsches Forschungszentrum f{\"u}r K{\"u}nstliche Intelligenz (DFKI)}
  \city{Kaiserslautern}
  \country{Germany}
}

\author{Alain Pagani}
\email{alain.pagani@dfki.de}
\orcid{0000-0002-5136-0837}
\affiliation{%
  \institution{Deutsches Forschungszentrum f{\"u}r K{\"u}nstliche Intelligenz (DFKI)}
  \city{Kaiserslautern}
  \country{Germany}
}

\renewcommand{\shortauthors}{Millerdurai et al.}

\ccsdesc[500]{Human-centered computing~Mixed / augmented reality}
\ccsdesc[300]{Computing methodologies~Motion capture}
\ccsdesc[100]{Computing methodologies~Neural networks}

\keywords{egocentric hand pose estimation, hand-arm reconstruction, monocular RGB, egocentric RGB, computer vision}

\begin{teaserfigure}
  \includegraphics[width=\textwidth]{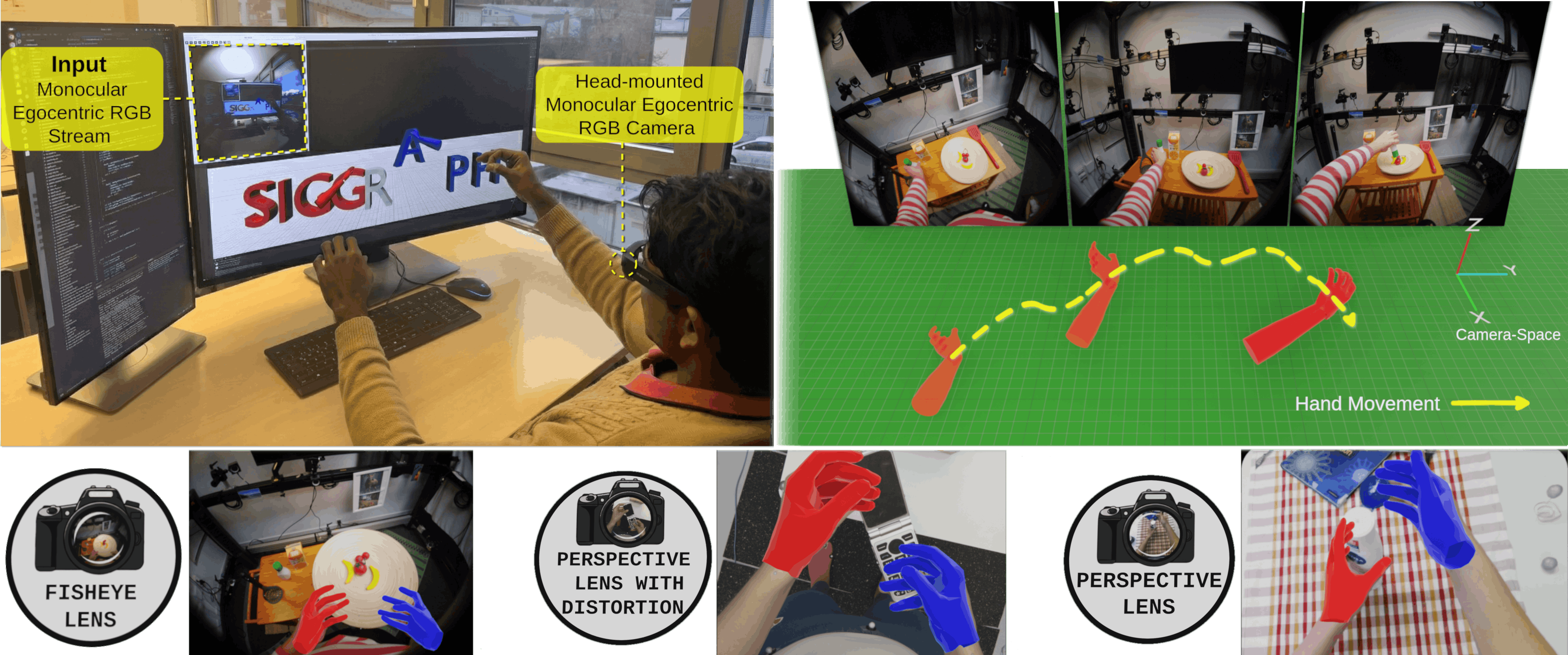}
  \caption{
\textbf{EgoForce reconstructs the absolute 3D pose and shape of the hands from the user’s viewpoint using a monocular RGB camera from Aria glasses (top left).}
With a unified framework, it supports diverse camera models while producing accurate 3D hand pose and shape (bottom), and recovers the absolute 3D hand position in the egocentric frame (top right), enabling metrically meaningful, viewpoint-consistent 3D tracking.  %
  }
  \Description{}
  \label{fig:teaser} 
\end{teaserfigure}

\begin{abstract}
{\fontsize{10}{10}\selectfont \textbf{Abstract}
}
\smallskip

Reconstructing the absolute 3D pose and shape of the hands from the user’s viewpoint using a single head-mounted camera is crucial for practical egocentric interaction in AR/VR, telepresence, and hand-centric manipulation tasks, where sensing must remain compact and unobtrusive.
While monocular RGB methods have made progress, they remain constrained by depth–scale ambiguity and struggle to generalize across the diverse optical configurations of head-mounted devices.
As a result, models typically require extensive training on device-specific datasets, which are costly and laborious to acquire.
This paper addresses these challenges by introducing \textsc{EgoForce}, a monocular 3D hand reconstruction framework that recovers robust, absolute 3D hand pose and its position from the user’s (camera-space) viewpoint.
\textsc{EgoForce} operates across fisheye, perspective, and distorted wide-FOV camera models using a single unified network.
Our approach combines a differentiable forearm representation that stabilizes hand pose, a unified arm–hand transformer that predicts both hand and forearm geometry from a single egocentric view, mitigating depth–scale ambiguity, and a ray space closed-form solver that enables absolute 3D pose recovery across diverse head-mounted camera models.
Experiments on three egocentric benchmarks show that \textsc{EgoForce} achieves state-of-the-art 3D accuracy, reducing camera-space MPJPE by up to \(28\%\) on the HOT3D dataset compared to prior methods and maintaining consistent performance across camera configurations.
For more details, visit the project page at \textcolor{sigteal}{\url{https://dfki-av.github.io/EgoForce}}.
\end{abstract}

\maketitle

\section{Introduction}
 
Following the shift from bulky head-mounted AR/VR systems toward compact wearable devices such as Project Aria (top-left of Fig.~\ref{fig:teaser}), devices increasingly rely on a lightweight perception stack built around a \emph{single} egocentric RGB camera.
Consequently, monocular egocentric 3D hand pose estimation becomes both essential and inherently challenging.
This capability is fundamental for applications such as onsite teleoperation and surgical training, where accurate hand motion must be recovered in the headset’s metric 3D frame from this single compact head-mounted camera.

Most existing monocular methods~\cite{HaMeR,WiLoR,MeshTransformer} estimate the 3D hand pose relative to a root joint (e.g., the wrist), recovering joint coordinates only up to an unknown translation and scale.
While this simplifies supervision (since only relative annotations are required), it cannot provide the hand’s absolute 3D position, which is essential for interaction-centric downstream tasks.
In contrast, we aim to recover the full 6-DoF hand pose, \ie camera-aware hand translation and orientation, directly in the headset’s metric coordinate frame, enabling plug-and-play integration with physics engines, grasp planners, collision-avoidance modules, and hand–object compositors without ad-hoc alignment, scale heuristics, or manual calibration.
However, achieving this from a single egocentric camera is challenging due to depth–scale ambiguity, frequent self-occlusions, and the strong distortions introduced by wide-FOV and fisheye optics~\cite{Millerdurai_EventEgo3D_2024,eventegoplusplus}.
These challenges are compounded by the wide variety of camera configurations used in egocentric AR/VR setups (\eg perspective, fisheye, cylindrical or spherical), whose differing projection models make it difficult to train a single monocular system that generalizes across optics.

\begin{figure*}[t]
  \centering
  \includegraphics[width=\linewidth]{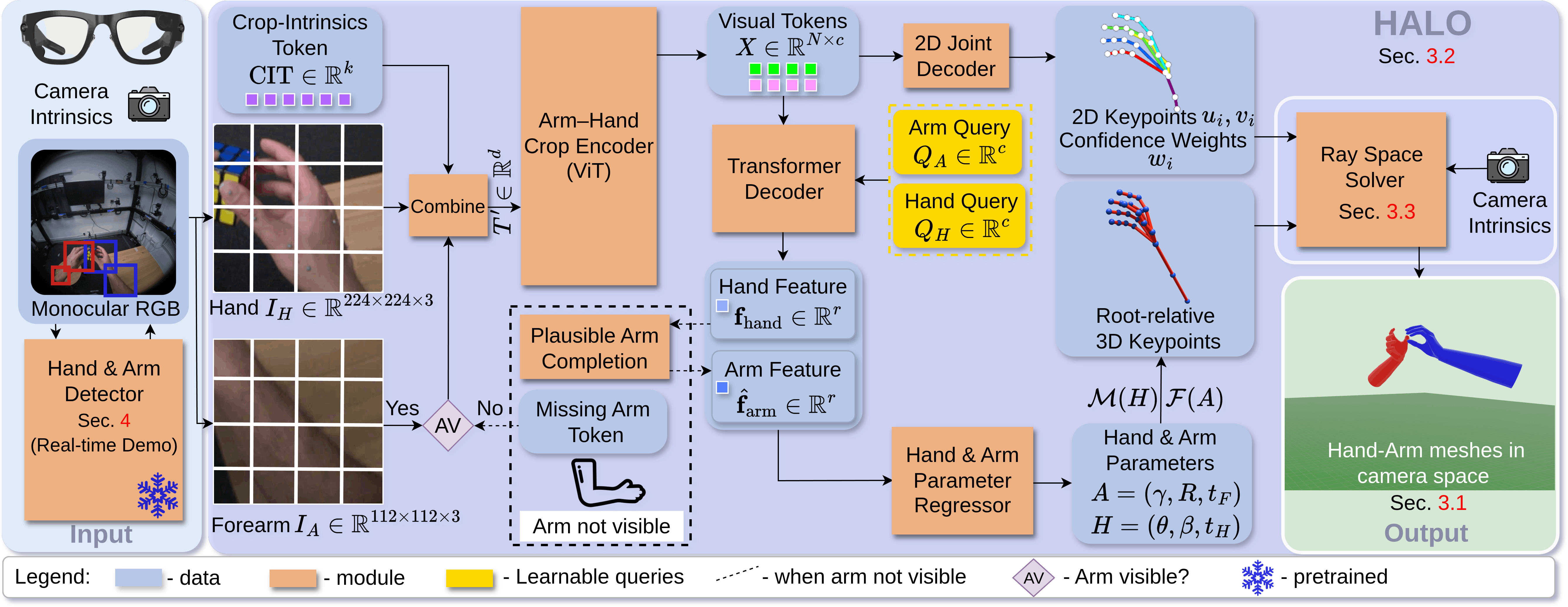} 
  \caption{%
    \textsc{EgoForce} processes a monocular egocentric RGB frame by extracting hand and forearm crops, tokenizing them, and conditioning the features on crop intrinsics (CIT).
    A transformer jointly infers hand–arm features to predict 2D keypoints (with confidences) and root-relative 3D hand and arm poses, which are lifted to camera-space meshes via the ray space solver.
    When the forearm is out of view, arm tokens are replaced with missing-arm tokens, and a hand-conditioned variational prior infers a plausible arm representation. 
    We apply this workflow independently to the left and right hand--forearm crops. 
    }
  \label{fig:pipeline}
\end{figure*}

To address these challenges, we introduce \textsc{EgoForce}, a camera-space 3D hand-pose estimation framework that jointly leverages hand and forearm imagery to recover absolute 3D hand pose from a single egocentric camera (Fig.~\ref{fig:pipeline}).
Our first key insight is that the forearm provides strong metric cues that help resolve monocular depth-scale ambiguity: by modeling the anthropometric\footnote{We use \emph{anthropometric} to refer to human body measurements and their statistical relationships (\eg correlations between limb-segment proportions).}
coupling between forearm and hand (\eg ANSUR~\citep{gordon1989ansur} reports strong correlations between forearm length and overall arm size) and their coupled movement, \textsc{EgoForce} reduces depth–scale ambiguity far more reliably than hand-only methods.
Our second key insight is a camera-model-agnostic ray space lifting formulation that operates on 2D joint observations formulated as rays rather than raw image coordinates, enabling a unified pipeline that generalizes across perspective, fisheye, and distorted wide-FOV optics.
Together, these components enable robust absolute (camera-space) 3D hand reconstruction from a single egocentric camera. To realize these insights, we propose the following contributions:
\begin{enumerate}[
leftmargin=14pt,
labelindent=-\parindent,
labelsep=0.5em,
partopsep=0pt,
topsep=5pt,
parsep=0pt
]
\item \textbf{H}and-\textbf{A}rm \textbf{L}atent-Shape \& \textbf{O}rientation (HALO), a unified regression architecture that jointly regresses hand and forearm pose with their shape proxies from monocular images.
\item A fully differentiable forearm representation that provides metric cues and improves absolute 3D hand-pose estimation through contextual arm–hand reasoning.
\item A cross-camera ray space solver that recovers absolute 3D hand- forearm placement across fisheye, perspective, and distorted wide-FOV optics, enabling deployment across diverse head-mounted optics.
\end{enumerate}

\section{Related Works}

\subsection{3D Hand Pose Estimation}
Monocular 3D hand pose estimation has progressed rapidly, but recovering \emph{absolute} (camera-space) hand pose from a single RGB image remains challenging due to depth ambiguity along with lens and crop-induced distortions.
Consequently, most methods~\cite{HaMeR,WiLoR,MeshTransformer} predict root-relative pose under weak perspective, discarding absolute hand position and often ignoring crop geometry effects~\cite{prakash20243d}.

Prior approaches regressing 3D hand poses in the camera space fall into several categories.
Single-stage regressors~\cite{Millerdurai_3DV2024} provide an efficient end-to-end formulation, but often struggle to generalize across cameras.
Root-depth regressors~\cite{Moon_2020_ECCV_I2L-MeshNet,Moon_2019_ICCV_3DMPPE} lift root-relative 3D poses to camera space by depth but rely on brittle anthropometric assumptions.
Methods using known intrinsics~\cite{GANeratedHands_CVPR2018,iqbal2018hand,zhou2020monocular} still require global scale estimation.
Implicit neural formulations~\cite{huang2023neural} regress camera-space joints via learned distance fields, but depend on accurate masks and manual tuning.
Registration-based pipelines~\cite{Park_2022_CVPR_HandOccNet,handdgp2024,MobRecon} decouple 2D detection and 3D lifting.
However, many predict a root-relative hand and perform post-hoc iterative registration~\cite{Park_2022_CVPR_HandOccNet,MobRecon}, limiting end-to-end camera-space reasoning.
HandDGP~\cite{handdgp2024} integrates differentiable registration but still relies on operating in a rectified or pinhole-style correspondence setting, making both correspondence learning and backpropagation through nonlinear projections fragile under extreme wide-FOV optics.
Our approach follows the registration paradigm but differs from previous registration-based methods in two key ways:
(1) Ray space alignment lifts the estimated 2D-3D correspondences directly in \emph{ray space}, i.e., using bearing vectors from the \emph{native calibrated projection model} (including fisheye/distorted wide-FOV), which removes dependence on a specific pixel coordinate system; while point-to-ray fitting is classic~\cite{ansar2003linear,pless2003using}, our novelty is integrating it as a stable lifting module and validating across camera models in egocentric hand tracking; and
(2) Our Crop Intrinsics Tokens (CIT) encode normalized crop intrinsics into transformer tokens to enable consistent geometric reasoning across lenses and crop configurations.
While multi-view methods such as UmeTrack~\cite{han2022umetrack} target camera-space hand tracking for VR headsets, we instead focus on monocular hand reconstruction from a single camera, a setting better suited to lightweight smart glasses.
Finally, some works estimate monocular world-space hand trajectories, but they rely on SLAM, explicit camera-motion estimation, multi-stage disentanglement frameworks \cite{HaWoR,Dynhamr}, or scale alignment \cite{HaPTICHands}, making them sensitive to drift and scale instabilities. 
In contrast, \textsc{EgoForce} predicts per-frame stable camera-space hand placements without SLAM, odometry, or manual scale calibration thanks to the ray space lifting module that directly solves point-to-ray constraints under the native camera model.

\subsection{Hand-Forearm Context Reasoning}
The forearm and hand are biomechanically coupled: forearm pose stabilizes orientation and provides a strong spatial prior that narrows the 3D space where the hand can plausibly be  \cite{liu2022spatial,lee2024enhancing}. 
Moreover, forearm size covaries with body scale, providing soft anthropometric priors that help resolve monocular depth–scale ambiguity~\cite{vukotic2023forearmstature,rostamzadeh2021handforearm}.
Existing approaches exploit this coupling in different ways.
\citet{tse2023spectral,lee2024enhancing} show that including the forearm region alongside the hand improves 3D hand pose estimation, either by regressing unified hand--forearm meshes or by leveraging forearm context for better hand accuracy.
However, directly enlarging the hand crop to include the forearm can dilute high-frequency hand details required for precise joint localization (CNNs). 
Even with tokenization in transformers, careful design is needed to avoid spurious outputs when the forearm is not visible.
To address these challenges, we propose a novel modular, crop‐based framework that (1) processes hand and forearm regions separately to preserve fine-grained hand geometry; and (2) fuses them via cross-attention to capture the kinematic structure. 
This design preserves fine-grained hand detail while still leveraging forearm information to reduce depth–scale ambiguity.
Moreover, when the forearm is not visible, a generative forearm prior infers a plausible arm orientation, maintaining continuity and realism in 3D hand–forearm motion.
Hereafter, ``arm'' also refers to the forearm for conciseness.

\section{The \textsc{EgoForce} Framework}
Fig.~\ref{fig:pipeline} overviews the proposed \textsc{EgoForce} approach for recovering camera-space 3D hand and forearm meshes from a single egocentric RGB frame. 
We define the hand and forearm models in Sec.~\ref{subsec:prelim}, describe our Hand-Arm Latent-Shape \& Orientation (HALO) architecture in Sec.~\ref{subsec:HALO} that predicts root-relative meshes and 3D joints, lift these predictions to camera space using our Ray Space Solver in Sec.~\ref{sec:RDLS_Solver_Main}, and detail supervision in Sec.~\ref{sec:losses}.

\subsection{Preliminaries}
\label{subsec:prelim}
\noindent \textbf{Hand Model.} 
We represent each hand using MANO~\cite{mano_hand}, with pose $\theta\in\mathbb{R}^{16\times 6}$ (6D rotations~\cite{zhou2018continuity}), shape $\beta\in\mathbb{R}^{10}$, and translation $t_H\in\mathbb{R}^3$. 
We denote hand parameters as $H=(\theta,\beta,t_H)$ and obtain mesh/joints via MANO operator $\mathcal{M}(H)$.

\noindent\textbf{Forearm model (FARM).}
We introduce a novel formulation for each forearm using FARM (ForeArm Representation Model) with shape $\gamma\in\mathbb{R}^{5}$, rotation $R\in\mathbb{R}^{6}$, and translation $t_F\in\mathbb{R}^{3}$.
We denote FARM parameters as $A=(\gamma,R,t_F)$ and obtain mesh/joints via FARM operator $\mathcal{F}(A)$.
For details regarding the construction and parameterization of FARM, please refer to the Sup.~Sec.~\ref{sec:FARM_repr}.

\begin{figure}[h]
  \centering
  \includegraphics[width=\linewidth]{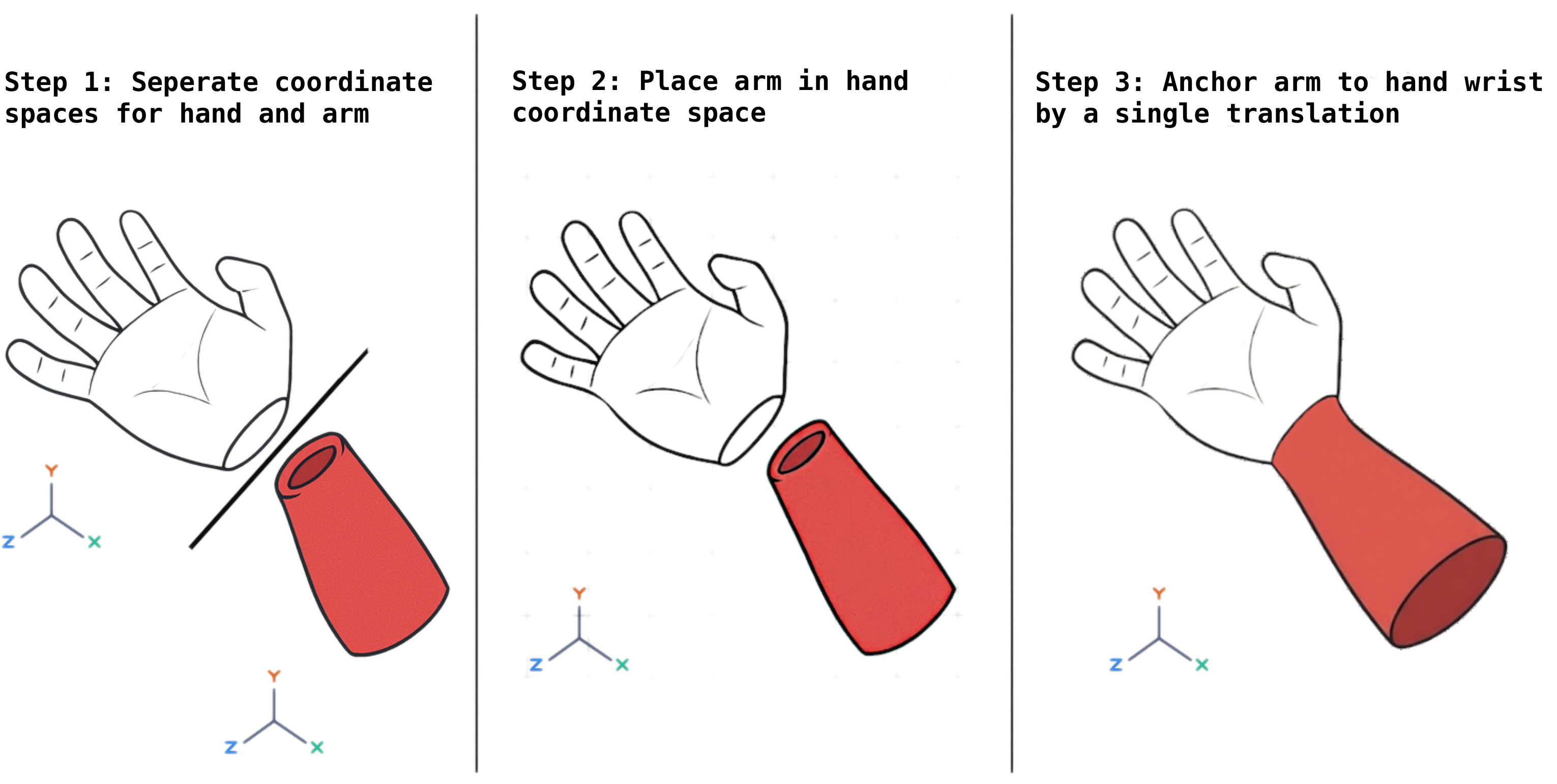} 
  \vspace{-10pt}
  \caption{
  \textbf{Unified hand–arm mesh.} We attach the FARM at the MANO wrist and apply a small elbow-direction offset to avoid overlap and ensure a clean, anatomically consistent connection.
    }
  \label{fig:limbmodel}
  \vspace{-1em}
\end{figure}

\noindent\textbf{Unified hand-arm mesh.}
We attach FARM to the MANO wrist by aligning FARM's wrist to the MANO wrist with a single translation in the MANO frame (see Sup.~Fig.~\ref{fig:limbmodel}). 
To avoid interpenetration, we apply a small elbow-direction offset (about $3\%$ of the elbow-to-wrist length) while preserving rotation, yielding a clean, anatomically consistent connection.
This procedure yields clean, non-intersecting geometry and ensures that the hand and forearm remain rigidly anchored in camera space, enabling stable and physically coherent camera-space estimates.

\subsection{\textbf{H}and-\textbf{A}rm \textbf{L}atent-Shape \& \textbf{O}rientation Architecture}
\label{subsec:HALO}

For each limb $\ell\in\{\text{left},\text{right}\}$, HALO takes a hand crop $\mathbf I_H^\ell\in\mathbb{R}^{224\times224\times3}$ and a forearm crop $\mathbf I_A^\ell\in\mathbb{R}^{112\times112\times3}$ as input.
Both crops are extracted from the original frame using their respective bounding boxes and then resized to the specified resolutions. 
HALO predicts: (i) 2D joints for hand and arm, (ii) per-joint confidences, (iii) MANO parameters, and (iv) FARM parameters.

\noindent\textbf{Arm–Hand Crop Encoder.}
We remove lens-specific nonlinear distortions from each image crop using undistortion mapping and split them into $N_H$ and $N_A$ patches.
These patches are linearly projected to $d$-dim tokens and augmented with positional encodings, yielding
$
\mathbf{T}_H \in \mathbb{R}^{N_H \times d}
\quad\text{and}\quad
\mathbf{T}_A \in \mathbb{R}^{N_A \times d}.
$
We encode crop-specific intrinsics (normalized crop geometry and viewing parameters; described in Sup.~Sec.~\ref{sec:CIT}) similar to \citet{prakash20243d} and produce \emph{Crop Intrinsics Token}
$\mathrm{CIT} \in \mathbb{R}^{k}$.
These $\mathrm{CIT}$ tokens are combined\footnote{
concatenate two vectors, reduce dimensionality via an MLP, then add residually to the original token; see Sup.~Fig.~\ref{fig:CaDBlock}.
} with every patch token yielding
$
\mathbf{T}_H' \in \mathbb{R}^{N_H \times d}
\quad\text{and}\quad
\mathbf{T}_A' \in \mathbb{R}^{N_A \times d}.
$
If the forearm is out of view, we replace $\mathbf{T}_A'$ with \texttt{[MASK]} tokens (i.e.,~missing-arm token). 
Finally, all the tokens are passed through a ViT backbone~\cite{dosovitskiy2020image} to obtain visual tokens of dimension $c$:
$
\mathbf{X} \in \mathbb{R}^{N \times c}
$
where $N = N_H + N_A$.

\noindent \textbf{Contextual Decoding of Hand–Arm Interactions.}
To extract the hand and arm features, we employ two sets of learnable queries---four hand queries (2D joints, global pose, hand shape, hand pose) and three arm queries (2D joints, arm shape, arm pose)---and, denote these query vectors collectively as $\mathbf Q_H\in\mathbb R^{c}$ and $\mathbf Q_A\in\mathbb R^{c}$, respectively. 
These queries are stacked to form the target sequence
$\mathbf Q_0=[\mathbf Q_H;\mathbf Q_A]\in\mathbb{R}^{2\times c},$
and decoded with a transformer decoder attending to \(\mathbf{X} \in \mathbb{R}^{N \times c}\).
After $L$ layers (with \(L=2\) in practice), we obtain
$\mathbf Q_L=[\mathbf f_{\text{hand}};\mathbf f_{\text{arm}}]\in\mathbb{R}^{2\times r}$, where $r$ is the decoded query dimension.
The decoder's self-attention provides cross-limb context, letting the model leverage arm cues to resolve hand occlusions (and vice versa) during hand–object interaction.
When the arm is not visible, updates to the arm query \(\mathbf{Q_A}\) are masked.
Finally, we combine $\mathrm{CIT}$ with $\mathbf f_{\text{hand}}$ and $\mathbf f_{\text{arm}}$    to reinforce geometric conditioning.

\noindent\textbf{Plausible Arm Completion.}  
In egocentric views, the user's arm is frequently outside the camera's FOV, making direct visual localization challenging or even impossible. 
Although our primary focus is accurate hand tracking, inferring a plausible full arm pose can greatly benefit downstream applications (\eg AR immersion or physics simulation). 
We, therefore, introduce a conditional variational prior~\cite{sohn2015learning} that models a latent arm code $\mathbf z_{\text{arm}}$ conditioned on hand features ${\mathbf f}_{\text{hand}}$.
When the arm is not visible (i.e., no arm bounding box is available), we sample $\mathbf z_{\text{arm}}$ from this prior and obtain a plausible arm feature  ${\mathbf f}^{\text{prior}}_{\text{arm}}$, which replaces the missing arm feature:
$$
\hat{\mathbf f}_{\text{arm}}
= \begin{cases}
{\mathbf f}_{\text{arm}}, & \text{if the arm is visible},\\
{\mathbf f}_{\text{arm}}^{\text{prior}}, & \text{otherwise}.
\end{cases}
$$
This leverages visual evidence when available and falls back to a learned hand-conditioned kinematic prior otherwise, yielding stable and realistic hand-arm configurations in egocentric scenarios. %

\noindent\textbf{2D Joint Decoder.}
We compute spatial attention between $\mathbf X \in \mathbb{R}^{N_H \times c}$ and the hand features to produce a hand-focused spatial map, and compute spatial attention between $\mathbf X \in \mathbb{R}^{N_A \times c}$ and the arm features to produce an arm-focused spatial map. 
We then decode heatmaps with a lightweight CNN as in ViTPose~\cite{xu2022vitpose} for $J_H=21$ hand joints and $J_A=3$ forearm joints.
Joint locations $(u_j,v_j)$ are then obtained by soft-argmax over the heatmaps.
Next, we bilinearly sample the spatial maps at these locations and use an MLP to predict per-joint confidence weights $w_j$ as in \citet{handdgp2024}.
These sampled features are also combined with the corresponding hand and arm features and passed to the parametric regressor.

\noindent\textbf{Parametric Regressor.}
The final stage of HALO takes the hand and arm features and applies two parametric regressors~\cite{kanazawa2018end} to regress the hand $H=(\theta,\beta,t_H=\mathbf 0)$ and arm $A=(\gamma,R,t_F=\mathbf 0)$ parameters.
The $t_H$ and $t_F$ are set to zero to obtain the root-relative parameters, and the camera-space positions are recovered using our Ray Space Solver.

\subsection{Ray Space Solver (RSS)} 
\label{sec:RDLS_Solver_Main} 

After obtaining root-relative joints, detected 2D joints and confidence weights from HALO, we estimate a single camera-space translation $\mathbf t\in\mathbb R^3$ shared by the hand--arm mesh. 
For every 2D keypoint $(u_i,v_i)$, we back-project it through the calibrated camera model to obtain a unit ray direction $\mathbf d_i$.
The camera-space 3D joint $\mathbf P_i(\mathbf t)=\mathbf t+\mathbf J_i$ must lie on its corresponding ray, \ie there exists an (unknown) depth $\lambda_i$ such that the point-on-ray constraint $\mathbf t+\mathbf J_i=\lambda_i\mathbf d_i$ holds.
We eliminate the unknown depths $\lambda_i$ by measuring only the component of each translated joint that is \emph{perpendicular} to its ray, using the orthogonal projector $\mathbf\Pi_i=\mathbf I-\mathbf d_i\mathbf d_i^\top$ onto the plane normal to the ray, thereby removing the depth component.
\begin{figure}[h]
  \centering
  \includegraphics[width=\linewidth]{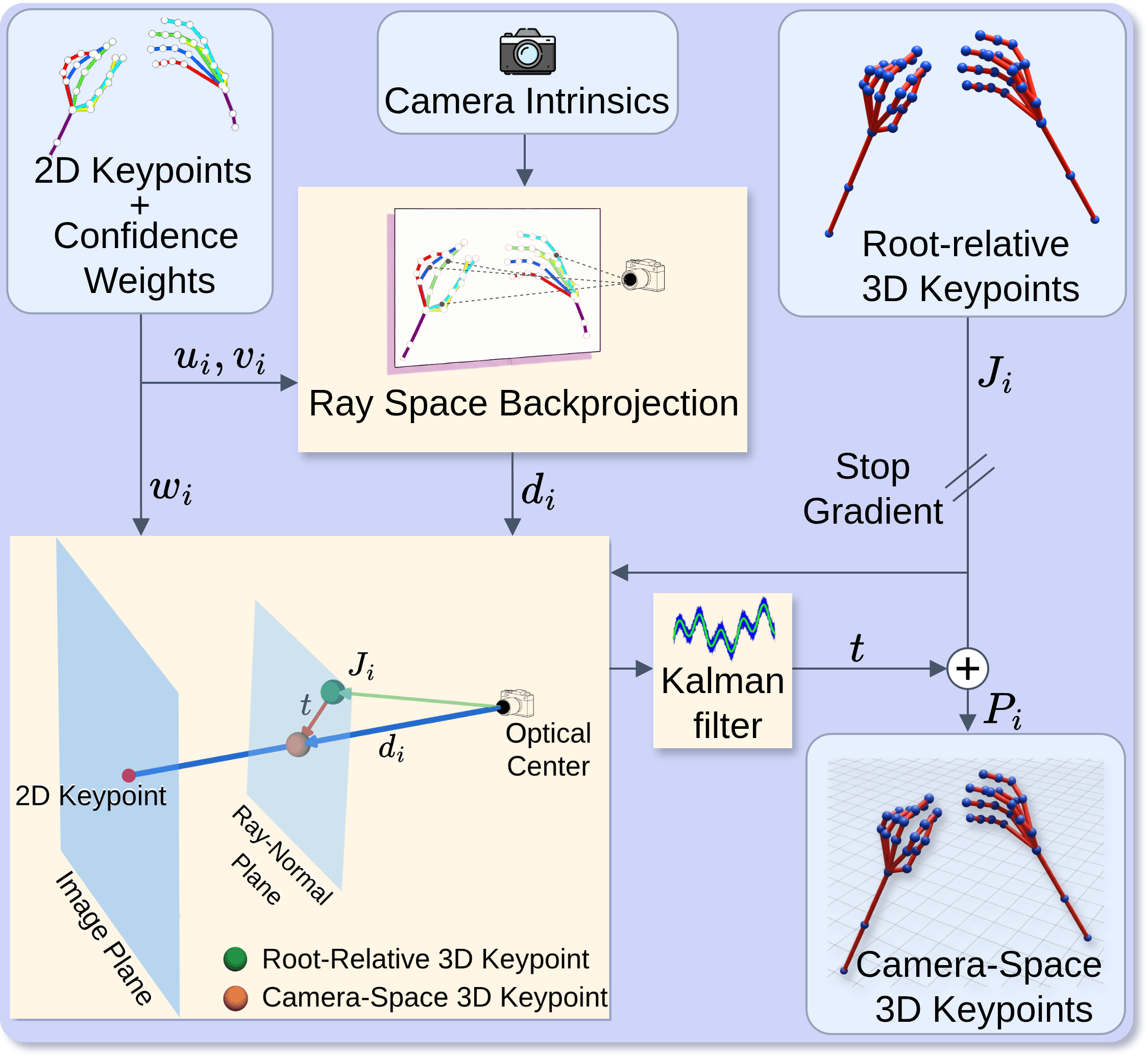} 
  \caption{
  \textbf{The Ray Space Solver} is a cross-camera (calibration-conditioned) module that recovers camera-space translation from 2D–3D correspondences, enabling deployment across devices with different optics.
    }
  \label{fig:Ray_Space_Solver}
\end{figure}
We estimate the shared translation by solving the confidence-weighted least-squares (closed-form) problem
\begin{equation}
\min_{\mathbf t}\;E(\mathbf t)
=\sum_{i=1}^{M} w_i\big\|\mathbf\Pi_i\mathbf P_i(\mathbf t)\big\|_2^2,
\label{eq:rss_ls}
\end{equation}
over all joints $i=1,\dots,M$, where $\mathbf\Pi_i\mathbf P_i(\mathbf t)$ is the depth-free point-to-ray residual and $\|\mathbf\Pi_i\mathbf P_i(\mathbf t)\|_2$ equals the perpendicular (\ie shortest) distance from the camera-space joint $\mathbf P_i(\mathbf t)$ to its ray (see Fig.~\ref{fig:Ray_Space_Solver}). 
To prevent occasional unstable camera-space fits from corrupting root-relative learning, we stop gradients through the solver from flowing back into the root-relative predictions.
Finally, we apply a Kalman filter to the per-frame translation estimates to improve stability under keypoint noise and occasional spurious solutions.
Since ray directions can be computed for any camera projection model, our Ray Space Solver generalizes to arbitrary calibrated cameras.
Full derivation of the solver is provided in Sup.~Sec.~\ref{sec:ray_depth_solve}, and Kalman filter details and hyperparameters are reported in Sup.~Sec.~\ref{supp:sec:kalman}.

\subsection{Loss Functions}\label{sec:losses} 

\noindent \textbf{2D Heatmap Loss}. 
Squared error between the predicted and ground‐truth 2D joint heatmaps for both hand and forearm:
\begin{equation}
\mathcal{L}_{\text{H}} =  \lambda_{H}^\mathcal{M} \frac{1}{N_{\mathcal{M}}}\sum_{i=1}^{N_{\mathcal{M}}} \lVert {H}^\mathcal{M}_{i} - \hat{{H}}^\mathcal{M}_{i} \rVert^2
+ \lambda_{H}^\mathcal{F} \frac{1}{N_{\mathcal{F}}}\sum_{i=1}^{N_{\mathcal{F}}} \lVert {H}^\mathcal{F}_{i} - \hat{{H}}^\mathcal{F}_{i}\rVert^2
,
\end{equation}
where $N_{\mathcal M}$ and $N_{\mathcal F}$ are the numbers of hand and forearm heatmaps, respectively; ${H}^\mathcal M_i,  \hat{H}^\mathcal M_i$ are the predicted and ground-truth 2D joint heatmap for the hand; 
${H}^\mathcal F_i,  \hat{H}^\mathcal F_i$ are the predicted and ground-truth 2D joint heatmap for the arm; and we set $\lambda_{H}^\mathcal{M}=20, \lambda_{H}^\mathcal{F}=100.$

\noindent \textbf{3D Joint Loss}. Squared error between the predicted and ground-truth \textit{root-relative} 3D joints:
\begin{equation}
\mathcal{L}_{\text{joints}} =  \lambda_{J}^\mathcal{M} \frac{1}{N_{\mathcal{M}}}\sum_{i=1}^{N_{\mathcal{M}}} \lVert {{J}}^\mathcal{M}_{i} - \hat{{J}}^\mathcal{M}_{i} \rVert^2
+ \lambda_{J}^\mathcal{F} \frac{1}{N_{\mathcal{F}}}\sum_{i=1}^{N_{\mathcal{F}}} \lVert {{J}}^\mathcal{F}_{i} - \hat{{J}}^\mathcal{F}_{i} \rVert^2
,
\end{equation}
where ${J}^\mathcal M_i,  \hat{J}^\mathcal M_i$ are the predicted and ground-truth joint coordinates; ${ J}^\mathcal F_i,  \hat{J}^\mathcal F_i$ are the predicted and ground-truth forearm joint coordinates; and we set $\lambda_{J}^\mathcal{M}=1$ and $\lambda_{J}^\mathcal{F}=5$. 
\noindent \textbf{MANO And FARM Losses}.
The hand pose $\theta$ and shape $\beta$, are penalized via an $\ell_2$ loss:
\begin{equation}
\mathcal L_{\mathrm{MANO}}
= \lambda_\theta \,\bigl\|\theta - \hat\theta\bigr\|^2
+ \lambda_\beta \,\bigl\|\beta - \hat\beta\bigr\|^2,
\end{equation}
where $\theta,\beta$ and $\hat\theta,\hat\beta$ are the predicted and ground-truth hand parameters, respectively, and we set
$\lambda_\theta=5,\;\lambda_\beta=0.01$.

Similarly, the forearm shape coefficients $\gamma$ and root rotation $R$ are supervised with
\begin{equation}
\mathcal L_{\mathrm{FARM}}
= \lambda_\gamma \,\bigl\|\gamma - \hat\gamma\bigr\|^2
+ \lambda_R \,\bigl\|R - \hat R\bigr\|^2,
\end{equation}
where $\gamma,\hat R$ and $\hat\gamma,\hat R$ are the predicted and ground-truth forearm parameters, and we choose
$\lambda_\gamma=0.5,\;\lambda_R=25$.

\noindent \textbf{Hand–Arm Relative Orientation Loss}.
To enforce consistency between hand and arm orientations, we first compute their relative rotations
$
R_{\mathrm{rel}}
= R_{\text{hand}}\,R_{\text{arm}}^\top,
\hat R_{\mathrm{rel}}
= \hat R_{\text{hand}}\,\hat R_{\text{arm}}^\top
$
where are $R$ the predicted rotations (converted from 6D representations) and $\hat R$ are the ground-truth rotations.
We then measure the mean angular misalignment using the SO(3) geodesic distance, $d_A(R_1,R_2)$ \cite{mahendran20173d} over the set of $\mathcal V$ frames where both the hand and the arm are present:
\begin{equation}
\mathcal L_{\mathrm{rel}}
=\lambda_{\mathrm{rel}} \frac{1}{|\mathcal V|}\sum_{i\in\mathcal V}
\bigl[d_A\bigl(R_{\mathrm{rel},i},\,\hat R_{\mathrm{rel},i}\bigr)\bigr]^2,
\end{equation}
and set $\lambda_{\mathrm{rel}}=0.5$.

\noindent \textbf{Forearm Prior Loss.} 
When the VAE prior is used to infill an invisible forearm, we compute the KL-divergence between the predicted prior and a standard Gaussian distribution:
\begin{equation}
\mathcal L_{\mathrm{prior}}
= \lambda_{\mathrm{prior}}~\mathrm{KL}\bigl(
   \mathcal N(\mu_{\mathrm{prior}},\,\sigma_{\mathrm{prior}}^2)
   \;\big\|\;
   \mathcal N(0, I)
 \bigr),
\end{equation}
and we set $\lambda_{\mathrm{prior}}=1$.

\noindent \textbf{Camera-Space 3D Joint Loss}.
Squared error between the predicted \textit{camera‐space} 3D joints and the ground‐truth 3D joint:
\begin{equation}
\mathcal{L}_{\text{cs}} =  \lambda_{P}^\mathcal{M} \frac{1}{N_{\mathcal{M}}}\sum_{i=1}^{N_{\mathcal{M}}} \lVert {{P}}^\mathcal{M}_{i} - \hat{{P}}^\mathcal{M}_{i} \rVert^2
+ \lambda_{P}^\mathcal{F} \frac{1}{N_{\mathcal{F}}}\sum_{i=1}^{N_{\mathcal{F}}} \lVert {{P}}^\mathcal{F}_{i} - \hat{{P}}^\mathcal{F}_{i}\rVert^2
,
\end{equation}
where ${P}^\mathcal M_i,  \hat P^\mathcal M_i$ are the predicted and ground-truth hand joint coordinates; ${ P}^\mathcal F_i,  \hat P^\mathcal F_i$ are the predicted and ground-truth forearm joint coordinates; and we set $\lambda_{P}^\mathcal{M}=0.001$ and $\lambda_{P}^\mathcal{F}=0.001$. 
\noindent \textbf{The Total Loss.} Overall, our total loss reads: 
\begin{multline}
\mathcal{L} = 
\mathcal{L}_{\text{H}} +  
\mathcal{L}_{\text{joints}}  +
\mathcal{L}_{\text{MANO}} + 
\mathcal{L}_{\text{FARM}} + 
\mathcal L_{\mathrm{rel}} + 
\mathcal L_{\mathrm{prior}} +
\mathcal{L}_{\text{cs}}.
\end{multline}
where the corresponding loss weights are included within each paragraph above. 
When FARM parameters are unavailable, we set $\mathcal{L}_{\mathrm{FARM}}$ and $\mathcal{L}_{\mathrm{rel}}$ to zero. In addition, hand and forearm keypoints that fall outside the image frame are masked out during supervision.

\section{Experimental Evaluation}

\noindent \textbf{Training and Evaluation Datasets.}
We train on six datasets: Re:InterHand~\cite{InterWild}, HandCO~\cite{zimmermann2021contrastive}, H2O~\cite{Kwon_2021_ICCV}, ARCTIC~\cite{fan2023arctic}, HO3D~\cite{hampali2020honnotate}, and HOT3D~\cite{banerjee2024introducing}. 
Re:InterHand, H2O, and ARCTIC include egocentric and exocentric views; HandCO and HO3D only exocentric; HOT3D only egocentric. 
In total, training data contain $3.67$M RGB images with MANO parameters and 3D joints. 
We evaluate mainly on egocentric H2O, HOT3D, and ARCTIC, and additionally on HO3D.
Additional details are in Supp.~Sec.~\ref {supp:subsec:dataset_splits}.

\noindent \textbf{FARM Generation.}
Since the datasets contain only MANO annotations, we generate FARM parameters and 3D arm joints per dataset.
For ARCTIC, we convert the provided SMPL-X meshes to FARM; for H2O, we triangulate multi-view imagery and fit FARM.
FARM recovery is infeasible for HO3D and HOT3D due to monocular depth ambiguity and frequent forearm occlusions, and for Re:InterHand and HandCO, the forearm is rarely visible, making FARM parameter recovery infeasible.
Additional details on FARM parameter generation are provided in Supp.~Sec.~\ref{supp:subsec:FARM_GEN}.

\noindent\textbf{Evaluation Metrics.}
We report (i) camera-space mean joint error (CS-MJE) from ~\citet{handdgp2024}, (ii) root-relative mean joint error (RS-MJE) from ~\citet{grauman2024ego}, (iii) Procrustes-aligned mean joint error (PS-MJE)~ from \citet{HaMeR}, and (iv) acceleration error (ACC) from ~\citet{Vibe}, reported as CS-ACC and RS-ACC.
Detailed definitions are provided in the Sup.~Sec.~\ref{supp:subsec:Evaluation_Metrics}.

\noindent\textbf{Evaluation Methodology.}
We train three camera-space baselines---MobRecon \cite{MobRecon}, HandOccNet \cite{Park_2022_CVPR_HandOccNet}, and HandDGP \cite{handdgp2024}---on our training data. 
As an additional baseline (\HaMerP), we use the pretrained HaMeR model \cite{HaMeR} and lift its root-relative predictions to camera space with DepthAnythingV2~\cite{depthanythingv2} and ground-truth intrinsics. 
We use the official implementations for MobRecon and HandOccNet and reimplement HandDGP (Sup.~Tab.~\ref{tab:hand_dgp_eval}).

\noindent \textbf{Implementation Details.}
We implement \textsc{EgoForce} in PyTorch \cite{paszke2019pytorch} and train with AdamW \cite{loshchilov2017decoupled}, batch size $27$, for $113$ epochs.
The transformer uses a learning rate of \(1\times10^{-5}\); all other modules use \(5\times10^{-4}\). 
Training on five NVIDIA H200 GPUs takes about four days. 
Further details are in Sup.~Sec.~\ref{supp:sec:extended_implementation}; all baselines use official hyperparameters.

\noindent \textbf{Live Demo and Runtime Performance.}
We demonstrate interactive performance in the supplementary video. 
Using the monocular fisheye stream from Aria glasses~\cite{engel2023projectarianewtool}, we detect hand-arm crops with RTMDet~\cite{lyu2022RTMDet}, regress camera-space hand and arm meshes with \textsc{EgoForce}, and stream them to \citet{Unity} for live rendering. 
The full pipeline runs at $\sim$14~FPS on an RTX~3090 for end-to-end two-hand tracking.

\subsection{Results}
\label{subsec:results}

Table \ref{tab:SOTA_results} reports quantitative comparisons against \HaMerP \cite{HaMeR}, MobRecon \cite{MobRecon}, HandOccNet \cite{Park_2022_CVPR_HandOccNet}, and HandDGP \cite{handdgp2024}.

\begin{figure}[h]
  \centering
  \includegraphics[width=\linewidth]{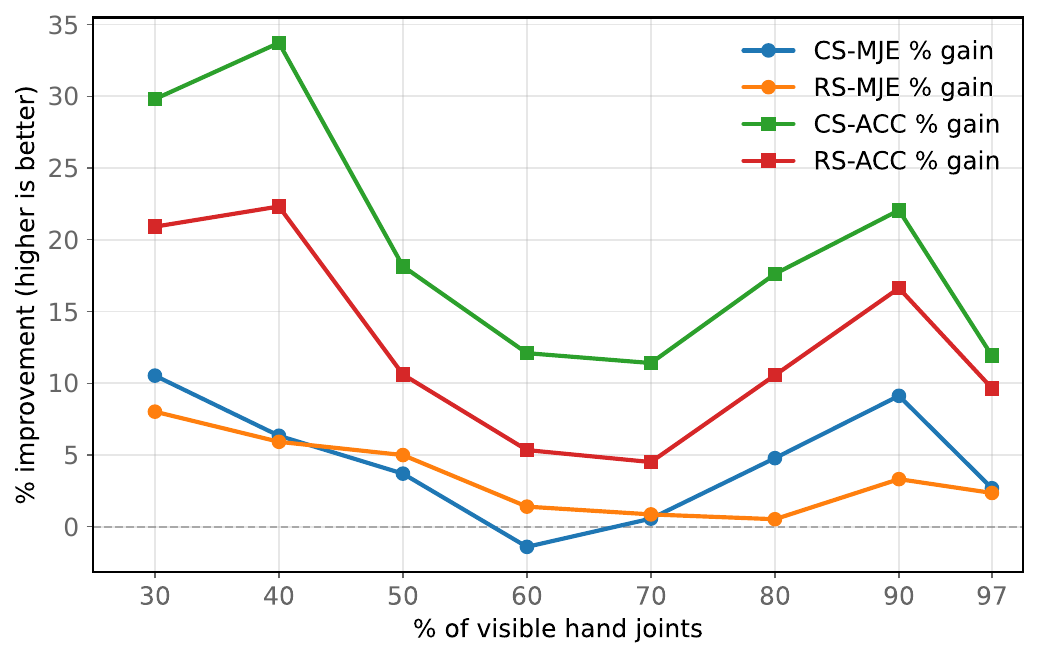} 
  \caption{
  \textbf{Influence of arm on hand-joint occlusion accuracy (ARCTIC dataset).} 
  Adding the arm consistently improves hand pose (RS-MJE), camera-space accuracy (CS-MJE), and temporal stability (RS-ACC, CS-ACC).
}
  \label{fig:ARCTIC_ARM_Hand_Vis}
\end{figure}

\noindent\textbf{ARCTIC.}
ARCTIC contains challenging egocentric scenarios with strong hand–object occlusions.
\HaMerP\ lifts 2D predictions to camera space using a monocular metric-depth estimator (DepthAnythingV2~\cite{depthanythingv2}), but depth estimation in the near field---where egocentric hands typically lie---is unreliable, leading to large CS-MJE. 
Prior work~\cite{HaWoR} reports similar degradation at extreme distances and proposes scene-based cues for improved stability, but such methods remain sensitive to motion blur, scene texture, illumination, and occlusion.
Two-stage pipelines such as MobRecon and HandOccNet achieve competitive articulation accuracy (PS-MJE) but struggle with camera-space localization.
HandDGP and \textsc{EgoForce} both use single-stage, feed-forward inference; HandDGP achieves good camera-space accuracy but weaker articulation, consistent with their report~\cite{handdgp2024}.
\textsc{EgoForce} achieves state-of-the-art performance in both articulation and camera-space accuracy.
As shown in Fig.~\ref{fig:ARCTIC_ARM_Hand_Vis}, incorporating arm context improves robustness under occlusion, yielding an overall $3\%$ improvement in RS-MJE and a $2.7\%$ improvement in CS-MJE. 
Temporal stability also improves notably, with CS-ACC reduced by $22\%$ and RS-ACC by $17\%$. 
The largest improvements in all metrics occur when hand-joint visibility is between 25-55$\%$ ($\approx$5-12 visible joints), which commonly arises during hand-object manipulation.

\begin{table*}[h]

\centering
\caption{
\textbf{Quantitative results on ARCTIC, HOT3D, H2O, and HO3D (in mm).}
\textcolor{siggold}{Gold} and \textcolor{sigbronze}{bronze} denote the best and second-best results, respectively. 
HO3D PS-MJE metrics of HandOccNet, \HaMerP, and MobRecon are from their official papers; HO3D CS-MJE metrics of MobRecon, HandOccNet, and HandDGP are from \citet{handdgp2024}.
}
\small
\begingroup
\setlength{\tabcolsep}{8.4pt}

\begin{tabular}{@{} 
    l 
    *{2}{S[table-format=3.2]} 
    c 
    *{2}{S[table-format=3.2]} 
    c
    *{2}{S[table-format=3.2]} 
    c 
    *{2}{S[table-format=3.2]} 
    @{}}
\toprule
Method 
  & \multicolumn{2}{c}{ARCTIC} 
  & 
  & \multicolumn{2}{c}{HOT3D} 
  & 
  & \multicolumn{2}{c}{H2O} 
  & 
  & \multicolumn{2}{c}{HO3D} \\
\cmidrule(lr){2-3}\cmidrule(lr){5-6}\cmidrule(lr){8-9}\cmidrule(lr){11-12}
  & {CS-MJE $\downarrow$} 
  & {PS-MJE $\downarrow$} 
  & 
  & {CS-MJE $\downarrow$} 
  & {PS-MJE $\downarrow$} 
  & 
  & {CS-MJE $\downarrow$} 
  & {PS-MJE $\downarrow$} 
  & 
  & {CS-MJE $\downarrow$} 
  & {PS-MJE $\downarrow$} \\
\midrule
\HaMerP                   & 2067.3  & 9.2  && 4493.7 & 8.3  && 631.6 & 6.3 && 561.5  & \bestMT{7.7}  \\
MobRecon                  & 81.5  & 9.6  && 116.3 & 8.0 && 49.1 & 6.2 && 121.7 & 9.2 \\
HandOccNet                & 256.3 & \bestMT{8.0}  && 284.8 & \bestMT{6.6} && 62.1 & \bestMT{5.3} && 156.4 & 9.1 \\
HandDGP                   & \second{51.7}  & 9.9  && \second{61.3} & 8.6  && \second{29.9} & 6.3                    && \second{50.3} & 9.3  \\
\midrule   
\textsc{EgoForce} (Ours)  & \bestMT{49.5}& \bestMT{8.0}  && \bestMT{43.9} & \bestMT{6.6}  && \bestMT{25.0} & \second{5.6} && \bestMT{49.5} & \second{9.0}  \\
\bottomrule
\end{tabular}
\endgroup
\label{tab:SOTA_results}
\end{table*}

\begin{figure}[h]
  \centering
  \includegraphics[width=\linewidth]{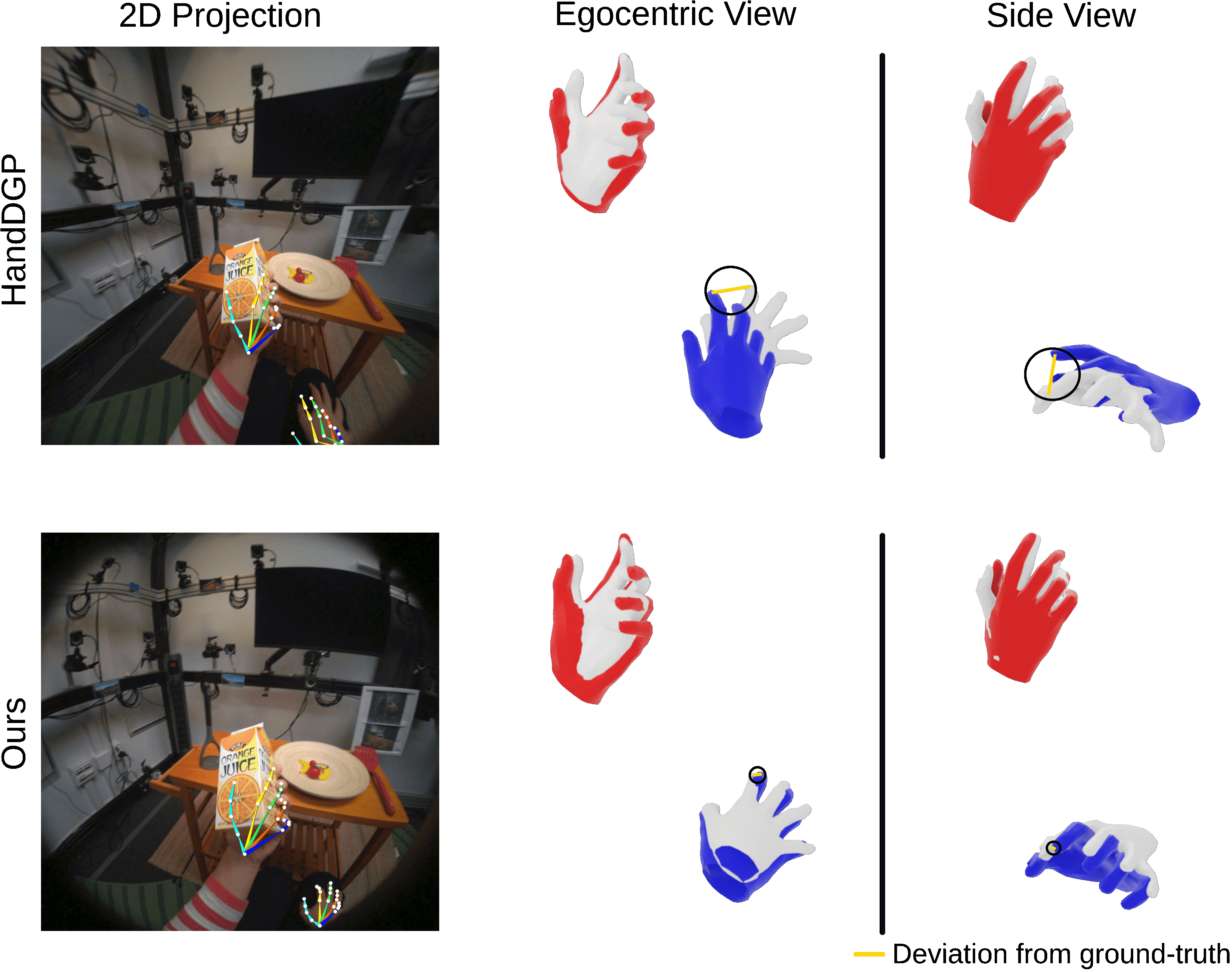} 
  \caption{
  \textbf{Camera-space results on HOT3D.}
    Left: egocentric input with the predicted 2D joint projections.
    Right: predicted meshes (left \textcolor{sigred}{red}, right \textcolor{sigblue}{blue}) and ground-truth meshes (gray) in camera space. 
    }
  \label{fig:HOT3D_Comparision}
\end{figure}

\begin{figure*}[h]
  \centering
  \includegraphics[width=\linewidth]{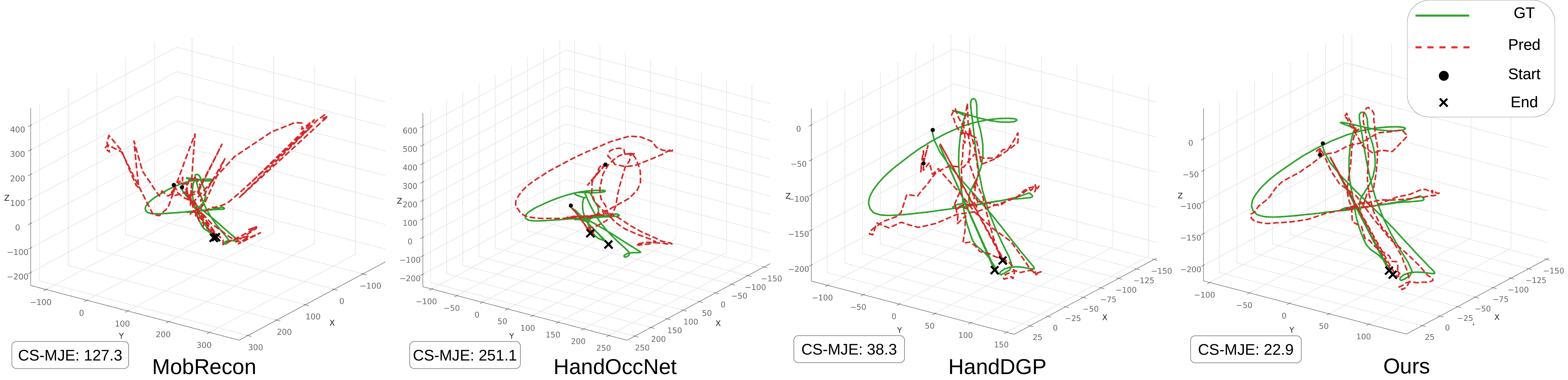} 
  \vspace{-2.1em}
  \caption{\textbf{Right-hand camera-space trajectory for a HOT3D sequence.} Our approach produces a more accurate hand trajectory in camera space, particularly along the depth (z-axis), compared to competing approaches. We visualize 160 frames from the sequence.
}
  \label{fig:HOT3D_Trajectory}
  \vspace{-1em}
\end{figure*}

\noindent \textbf{HOT3D.}
HOT3D is challenging due to (1) large-range hand motion during object interaction and (2) severe fisheye distortion combined with wide-FOV imagery, which amplifies depth ambiguity.
\textsc{EgoForce} achieves the highest accuracy across all methods, reducing CS-MJE by 28\% relative to HandDGP.
HandDGP relies on a pinhole-based 2D–3D correspondence formulation, making it sensitive to distortion near the image periphery, where fisheye effects dominate.
As illustrated in Fig.~\ref{fig:HOT3D_Comparision}, our method maintains accurate 3D reconstructions even when the hand moves toward the periphery, whereas HandDGP can deviate significantly despite plausible 2D projections.
Furthermore, \textsc{EgoForce} also produces the most accurate camera-space trajectories (Sup.~Fig.~\ref{fig:HOT3D_Trajectory}) for sequences, with both the start and end points closely matching the ground truth.

\begin{figure*}[t]
    \centering
    \includegraphics[width=\linewidth]{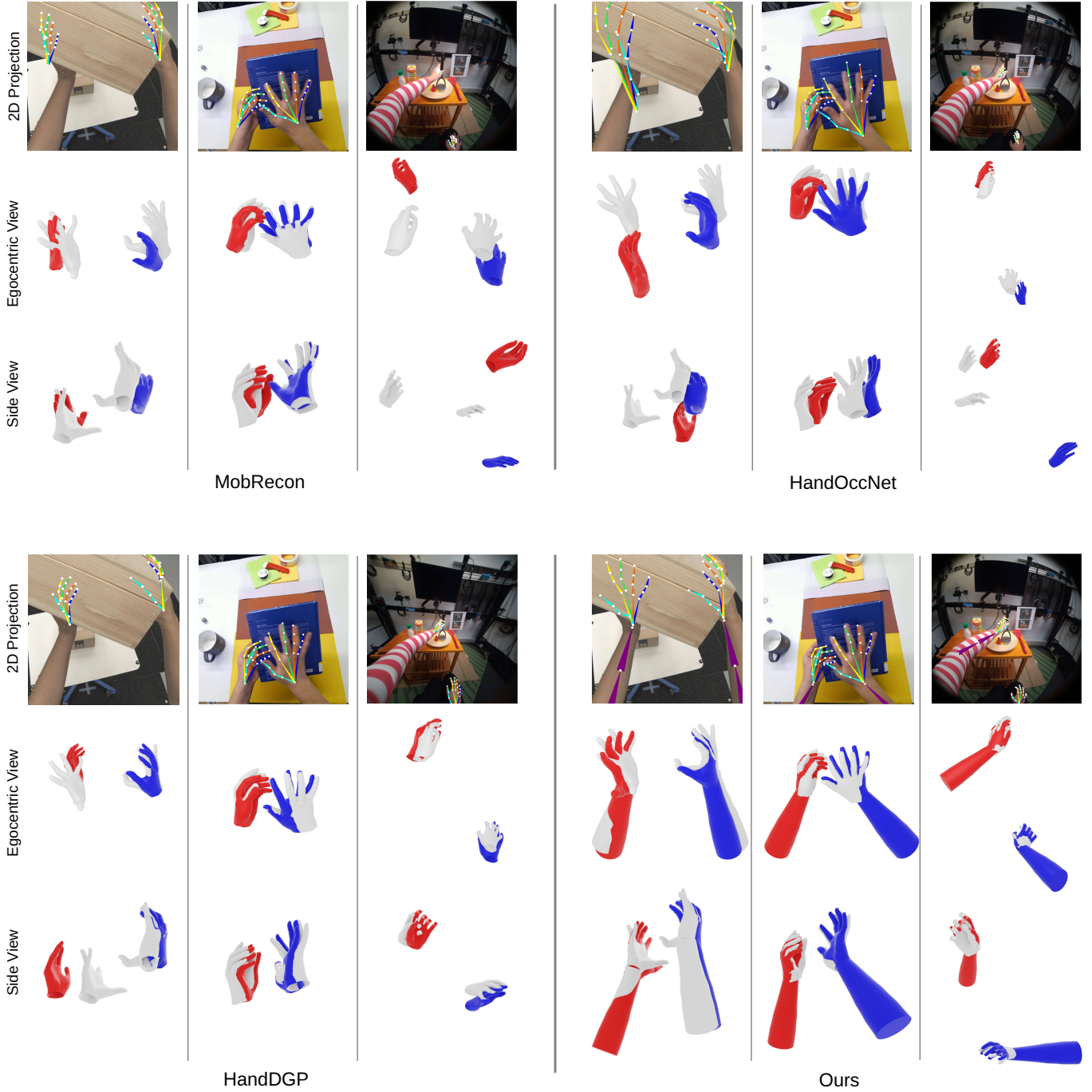}
    \vspace{-1.4em}
    \caption{\textbf{Qualitative camera-space results on egocentric datasets.}
    We compare our method against three state-of-the-art camera-space 3D hand pose methods on three datasets with widely different camera intrinsics.
    Predicted left and right limb meshes are shown in \textcolor{sigred}{red} and \textcolor{sigblue}{blue}, respectively, with ground truth highlighted in gray.
    }
    \label{fig:qualitative_comparision_datasets}
    \vspace{-1em}
\end{figure*}

\noindent \textbf{H2O.}
H2O contains highly dexterous hand–object interactions with pre-rectified (undistorted) images.
We outperform all competing methods in CS-MJE and achieve strong results in PS-MJE.
As shown in Tab.~\ref{tab:H2O_results}, it also attains the lowest CS-ACC, indicating the best temporal smoothness, and consistently recovers accurate hand orientation and articulation, as reflected by RS-MJE.
As illustrated in Fig.~\ref{fig:qualitative_comparision_datasets}, robust hand orientation estimates can be observed even under challenging object-interaction scenarios (e.g., holding a book).

\begin{table}[!t]
\caption{\textbf{Camera-space acceleration error (CS-ACC, in m/s$^2$) and root-relative hand pose error (RS-MJE, in mm) on the H2O dataset.}}
\centering
\small
\setlength{\tabcolsep}{6pt}
\begin{tabular}{@{\hspace{6pt}}l@{\hspace{50pt}}cc@{}}
\toprule
Method & CS-ACC $\downarrow$ & RS-MJE $\downarrow$ \\
\midrule
\HaMerP          & 55.9  & 19.0  \\
MobRecon         & 21.7  & 22.6  \\
HandOccNet       & 11.7 & 17.9 \\
HandDGP          & 8.5 & 17.3  \\
\midrule
\textsc{EgoForce} (Ours) & \best{5.5} & \best{14.8} \\
\bottomrule
\end{tabular}
\label{tab:H2O_results}
\end{table}

\noindent \textbf{HO3D.}
Although HO3D is captured from an external viewpoint rather than egocentric, we report results for completeness.
Our method achieves the lowest CS-MJE, surpassing the previous state of the art, HandDGP.
However, our PA-MJE (9.0$~mm$) is slightly higher than \HaMerP’s 7.7$~mm$, likely due to their use of extensive in-the-wild 2D and diverse 3D training data, whereas our training is restricted to 3D-annotated datasets.

\noindent \textbf{Comparison to Other Methods.}
EgoForce achieves lower CS-MJE on ARCTIC (55.1→49.5$~mm$) and lower PS-MJE on ARCTIC (14.7→8.0$~mm$), HOT3D (12.1→6.6$~mm$), and H2O (11.1→5.6$~mm$) against \citet{han2022umetrack}. EgoForce also outperforms \citet{HaWoR}, reducing CS-MJE from 319.9→49.5$~mm$ on ARCTIC and 72.5→25.0$~mm$ on H2O. See Sup. Sec.~\ref{subsec:additional_SOTAs} for detailed comparisons.

\subsection{Ablations}
\begin{figure}[t]
    \centering
    \includegraphics[width=\linewidth]{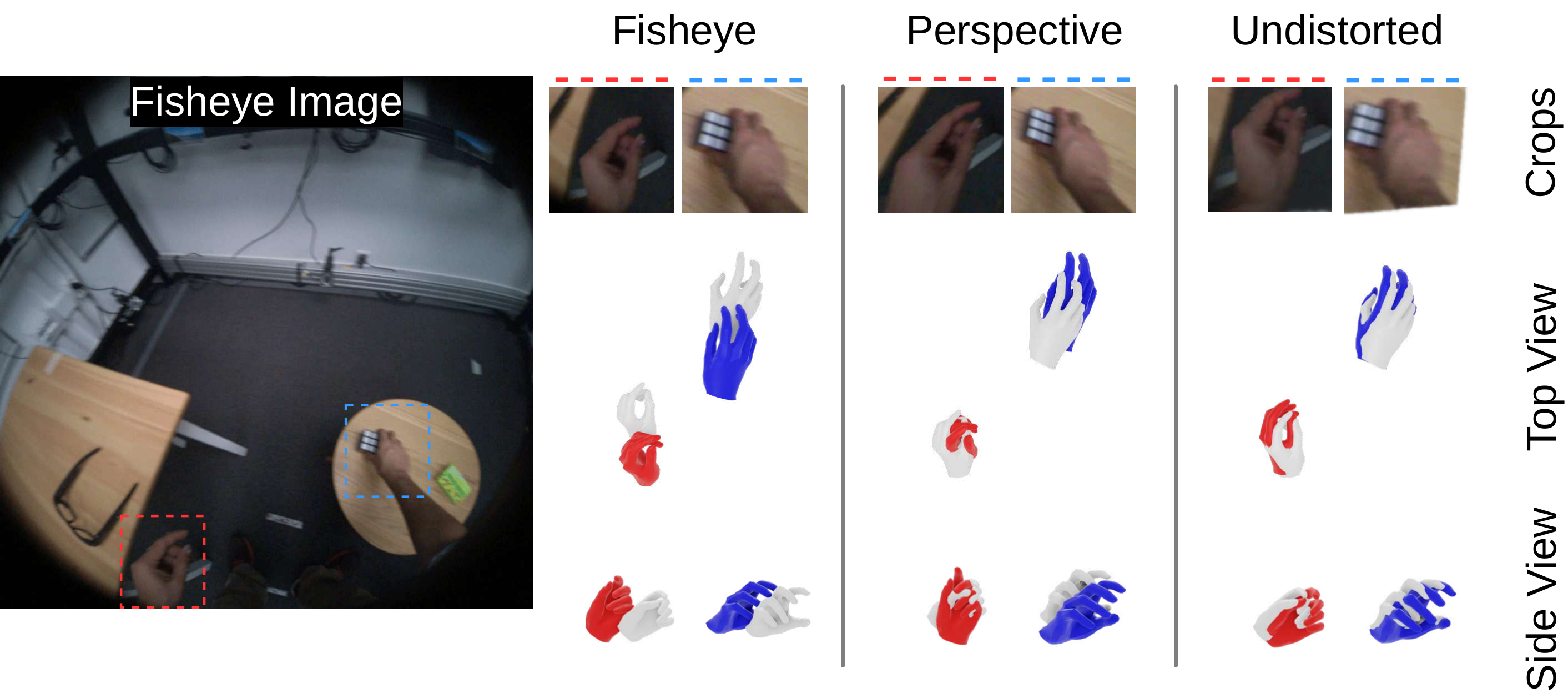}
    \vspace{-2.4em}
    \caption{\textbf{Influence of undistortion on input crops.}
    Direct hand-arm crops from the raw fisheye image lead to large errors as fisheye pixels correspond to highly non-linear viewing rays.
    Rectifying the full frame to a single perspective view reduces distortion but introduces strong peripheral warping and resampling artifacts that amplify localization noise.
    In contrast, lens-model undistortion preserves the correct pixel-to-ray geometry, yielding the most accurate camera-space reconstruction, especially near the image periphery.
    }
    \label{fig:ablation_w_distortion}
\end{figure}

\begin{figure}[t]
    \centering
    \includegraphics[width=\linewidth]{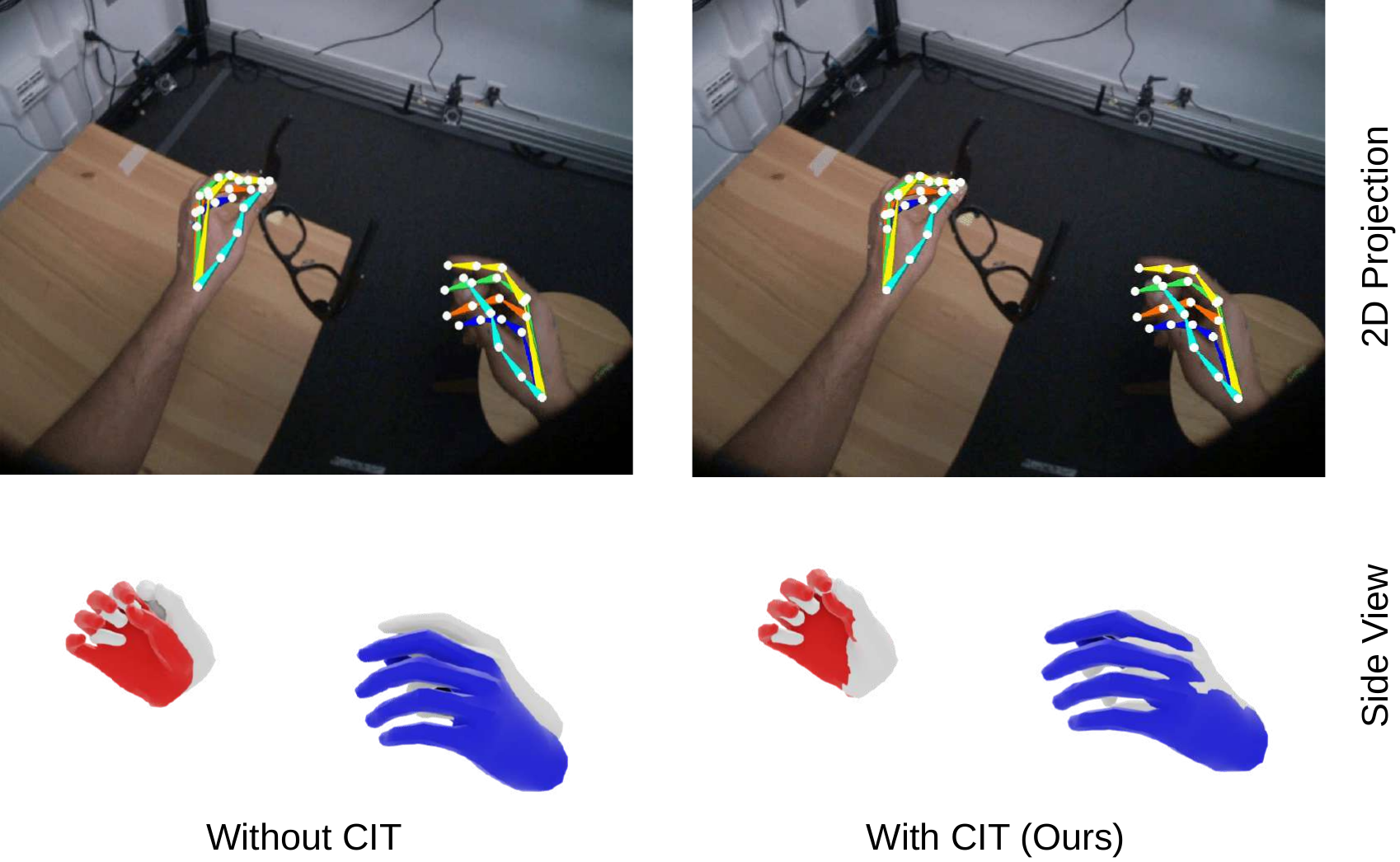}
    \vspace{-2.4em}
    \caption{\textbf{Influence of Crop Intrinsics Tokens (CIT).}
    CIT encodes crop-specific intrinsics as tokens for the hand-arm crop inputs feed to the transformer. 
    This enables explicit local camera-geometry reasoning and reduces camera-space mesh error, leading to closer alignment with ground truth.
    }
    \label{fig:ablation_w_cit}
\end{figure}

We isolate the key components of our method:

\noindent
\textbf{Camera geometry modeling.} 
The ablation on HOT3D (Tab.~\ref{tab:ablation_CIT}) highlights the importance of accurate camera modeling under extreme fisheye optics.
Undistortion alone provides the largest single gain (A→D), reducing CS-MJE from 123.4→48.7$~mm$ ($60.5\%\downarrow$) and RS-MJE from 53.1→19.7$~mm$ ($62.9\%\downarrow$). 
Introducing the Crop Intrinsics Token (A→B) without undistortion also helps ($37.9\%\downarrow$ on CS-MJE), but 
combining undistortion and crop-specific intrinsic conditioning (D→E) yields the best results: 45.8$~mm$ CS-MJE and 18.9$~mm$ RS-MJE.
In contrast, full-frame rectification (B→C) degrades performance due to the peripheral unwarping. 
Overall, explicit intrinsic modeling reduces CS-MJE by $62.9\%$ and RS-MJE by $64.4\%$ over the baseline, showing that camera geometry handling is crucial for accurate camera-space hand pose in fisheye imagery.
Please refer to Sup.~Fig.~\ref{fig:ablation_w_distortion}, Fig.~\ref{fig:ablation_w_cit}, and the video for qualitative examples illustrating the impact of undistortion and CIT.

\begin{table}[!h]
\caption{
\textbf{Ablation of Camera Geometry Modeling.}
CS-MJE, RS-MJE, and PS-MJE on HOT3D (in mm).
“CIT” = Crop Intrinsics Token; “Rect.” = Rectification; “Un.D” = Undistortion.
}
\centering
\small
\setlength{\tabcolsep}{4pt}
\renewcommand{\arraystretch}{1.1}
\begin{tabular}{@{} 
  l    
  c    
  c    
  c    
  S[table-format=2.2] 
  S[table-format=2.2] 
  S[table-format=1.2] 
@{}}
\toprule
Config. & CIT & Rect. & Un.D & {CS-MJE $\downarrow$} & {RS-MJE $\downarrow$} & {PS-MJE $\downarrow$} \\
\midrule
(A) & \xmark & \xmark & \xmark & 123.4 & 53.1 & 7.6 \\
(B) & \cmark & \xmark & \xmark & 76.6 & 29.0 & 6.8 \\
(C) & \cmark & \cmark & \xmark & 77.3 & 34.2 &  8.0 \\
(D) & \xmark & \xmark & \cmark & 48.7 & 19.7 & 6.6 \\
(E) & \cmark & \xmark & \cmark & \best{45.8} & \best{18.9} & \best{6.6}\\
\bottomrule
\end{tabular}
\label{tab:ablation_CIT}
\end{table}

\begin{figure}[t]
    \centering
    \includegraphics[width=\linewidth]{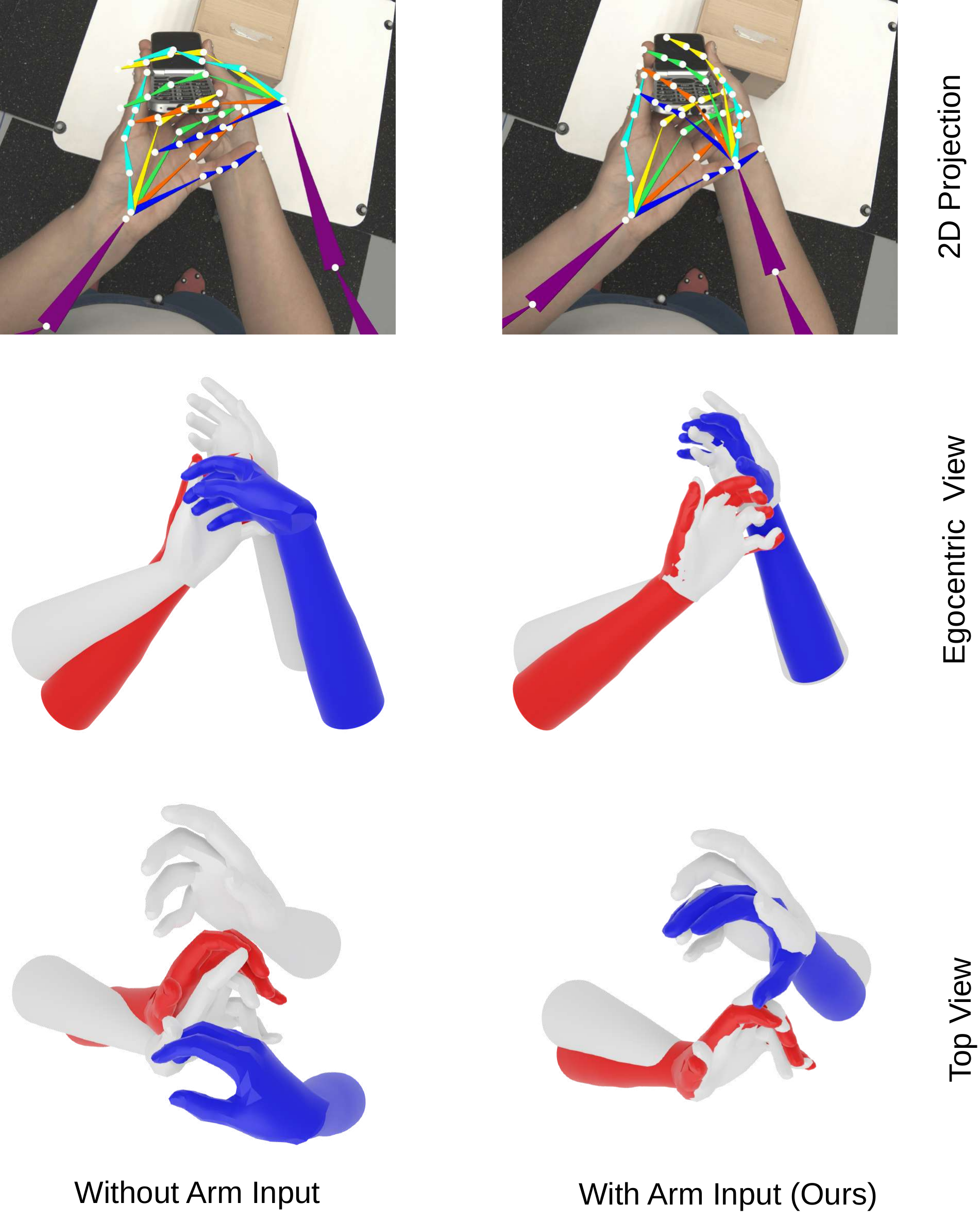}
    \vspace{-2.4em}
    \caption{\textbf{Influence of arm input.}
    Providing the arm crop as an input to the network improves hand pose accuracy.
    In this example, the right hand is strongly occluded by the phone and the other hand, yet the model recovers a plausible 3D pose, with accurate 2D joint reprojections and a hand-arm mesh closely aligned to ground truth.
}
    \label{fig:With_and_without_arm}
\end{figure}

\begin{figure}[t]
    \centering
    \includegraphics[width=\linewidth]{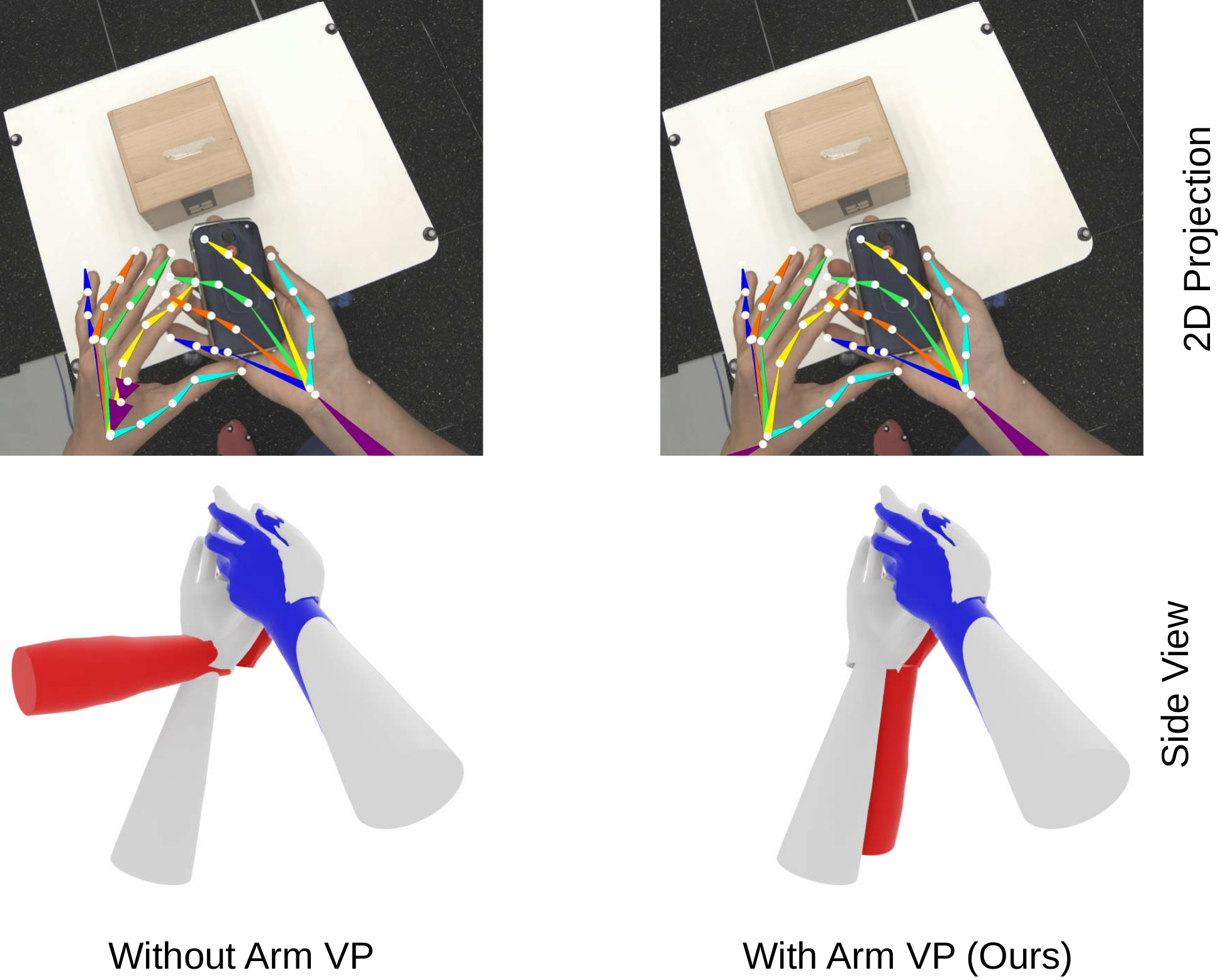}
    \vspace{-2.4em}
    \caption{\textbf{Influence of the variational arm prior.}
    Without the variational prior, the forearm is often mislocalized when it is heavily occluded.
    With the prior, the model infers a plausible forearm pose; in this example, the forearm is entirely out of view, yet the predicted position and orientation closely match ground truth.
    }    
\label{fig:With_and_without_VAE}
\end{figure}

\noindent
\textbf{Incorporating arm context.} 
We analyze the effect of arm context on ARCTIC for frames where the arm is visible and where it is not (Tab.~\ref{tab:arm-cvae-arctic}).
When the arm is visible, adding the arm crop improves hand performance, reducing CS-ACC from 19.3→15.2$~m/s^2$ ($21.2\%\downarrow$) and RS-MJE from 18.5→18.0$~mm$ ($2.7\%\downarrow$). 
Arm accuracy also improves (RS-MJE 20.4→17.0$~mm$), although arm CS-ACC slightly increases (20.7→22.72$~m/s^2$).
This suggests that while the model’s prior favors smooth but mean arm configurations and providing true arm evidence triggers more expressive articulation at the cost of a small reduction in temporal smoothness.
When the arm is not visible, introducing our hand-conditioned variational prior significantly improves arm estimates:
RS-MJE drops from 28.7→12.8$~mm$ ($55.4\%\downarrow$) and CS-ACC from 20.6→18.4$~m/s^2$ ($10.7\%\downarrow$), while hand performance remains unchanged. 
Overall, explicit arm conditioning benefits the hand when the arm is observed, and the learned arm prior is crucial for plausible arm recovery under occlusion, highlighting the importance of arm context for robust egocentric hand–arm pose estimation.
Please refer to Fig.~\ref{fig:With_and_without_arm}, Fig.~\ref{fig:With_and_without_VAE}, and the supplementary video for qualitative illustrations.

\begin{table}[!h]
  \caption{
  \textbf{Ablation of Arm.} 
  CS-ACC (in \si{m/s^2}) and RS-MJE (in mm) on the ARCTIC dataset.``VP'' = Variational Prior; ``Inp.'' = Input Crop;
  ``Vis.~F'' = Arm-visible frames; ``Invis.~F'' = Arm-invisible frames.
  }
  \centering
  \small
  \begingroup
  \setlength{\tabcolsep}{3pt}
  \begin{tabular}{@{}>{\centering\arraybackslash}m{4mm}
      l
      *{2}{S[table-format=2.2]}
      *{2}{S[table-format=2.2]}
      @{}} 
    \toprule
    & Method & \multicolumn{2}{c}{Hand} & \multicolumn{2}{c}{Arm} \\
    \cmidrule(lr){3-4}\cmidrule(lr){5-6}
    &        & {CS-ACC $\downarrow$} & {RS-MJE $\downarrow$} & {CS-ACC $\downarrow$} & {RS-MJE $\downarrow$} \\
    \midrule
    \multirow{2}{*}{\rotatebox{90}{\footnotesize Vis.~F}}
      & w/o Arm Inp.        & 19.3 & 18.5 & \best{20.7} & 20.4 \\
      & w/  Arm Inp. (Ours) & \hspace{-4pt} \best{15.2} & \best{18.0} & 22.7 & \best{17.0} \\
    \specialrule{.4pt}{2pt}{4pt}
    \multirow{2}{*}{\rotatebox{90}{\footnotesize Invis.~F}}
      & w/o Arm VP        & 10.5  & 6.0  & 20.6  & 28.7 \\
      & w/  Arm VP (Ours) & \best{10.5} & \best{6.0} & \best{18.4} & \best{12.8} \\
    \specialrule{.7pt}{4pt}{2pt}
  \end{tabular}
  \endgroup
  \label{tab:arm-cvae-arctic}
\end{table}

\noindent
\textbf{Depth--scale mitigation and hand-scale stability.}
Tab.~\ref{tab:depth_scale} quantitatively shows that forearm cues mitigate monocular depth--scale ambiguity. 
In particular, arm input reduces hand-scale error from 4.7 to 2.7$~mm$ when the hand lies 200--300$~mm$ from the camera (near field). 
In addition, frame-wise hand-scale variation for the sequences in datasets remains low, at 4$~mm$ on HOT3D and 2$~mm$ on ARCTIC, across 5 unseen hand sizes. 
See Sup. Sec.~\ref{subsec:depth_scale}.

\noindent
\textbf{Calibration-mismatch robustness.}
As shown in Fig.~\ref{fig:intrinsics_robustness}, EgoForce remains stable under intrinsic errors. 
On HOT3D, CS-MJE improves from 43.9 to 39.3$~mm$ at $50\%$ intrinsic noise, despite a camera-geometry error of 25.3$~mm$, and degrades gracefully only under large mismatches ($> 150\%$). 
See Sup. Sec.~\ref{subsec:intrinsics_robustness}.

\section{Limitations}
Our method is not without limitations.
It relies on calibrated 3D datasets for training, preventing the use of large 2D hand datasets common in root-relative methods~\cite{HaMeR,WiLoR} and limiting generalization to in-the-wild imagery. 
It also remains sensitive to camera intrinsics (See Fig.~\ref{fig:intrinsics_robustness} and Tab.~\ref{tab:camera_space_lifting} of Supplement).
Extended discussion of limitations are in Sup.~Sec.~\ref{supp:sec:extended_limitations}.

\section{Conclusion}

We introduce \textsc{EgoForce}, a monocular egocentric method for absolute camera-space 3D hand pose that leverages forearm context and camera-model--aware ray-space lifting.
Across three egocentric benchmarks, it delivers higher camera-space accuracy and stable temporal predictions, even with occlusions caused by hand--hand and hand--object interactions, and remains effective across both perspective and fisheye optics.
Ablations show that wide-FOV tracking benefits strongly from \emph{explicit camera geometry}: modeling distortion and conditioning on crop-aware intrinsics consistently improve performance.
We hope this work motivates future egocentric hand tracking systems to integrate forearm context and ray-based geometric constraints, especially as AR/VR hardware continues to shift toward compact, wide-FOV wearable cameras.

\section*{ACKNOWLEDGMENTS}
This work was partially funded by the Horizon Europe programme under the projects dAIEDGE, Grant Agreement No. 101120726, and IRIS-XR, Grant Agreement No. 101298672.
The authors thank the anonymous reviewers for their valuable feedback.

\clearpage

\begin{figure*}[t]
    \centering
    \includegraphics[width=\linewidth]{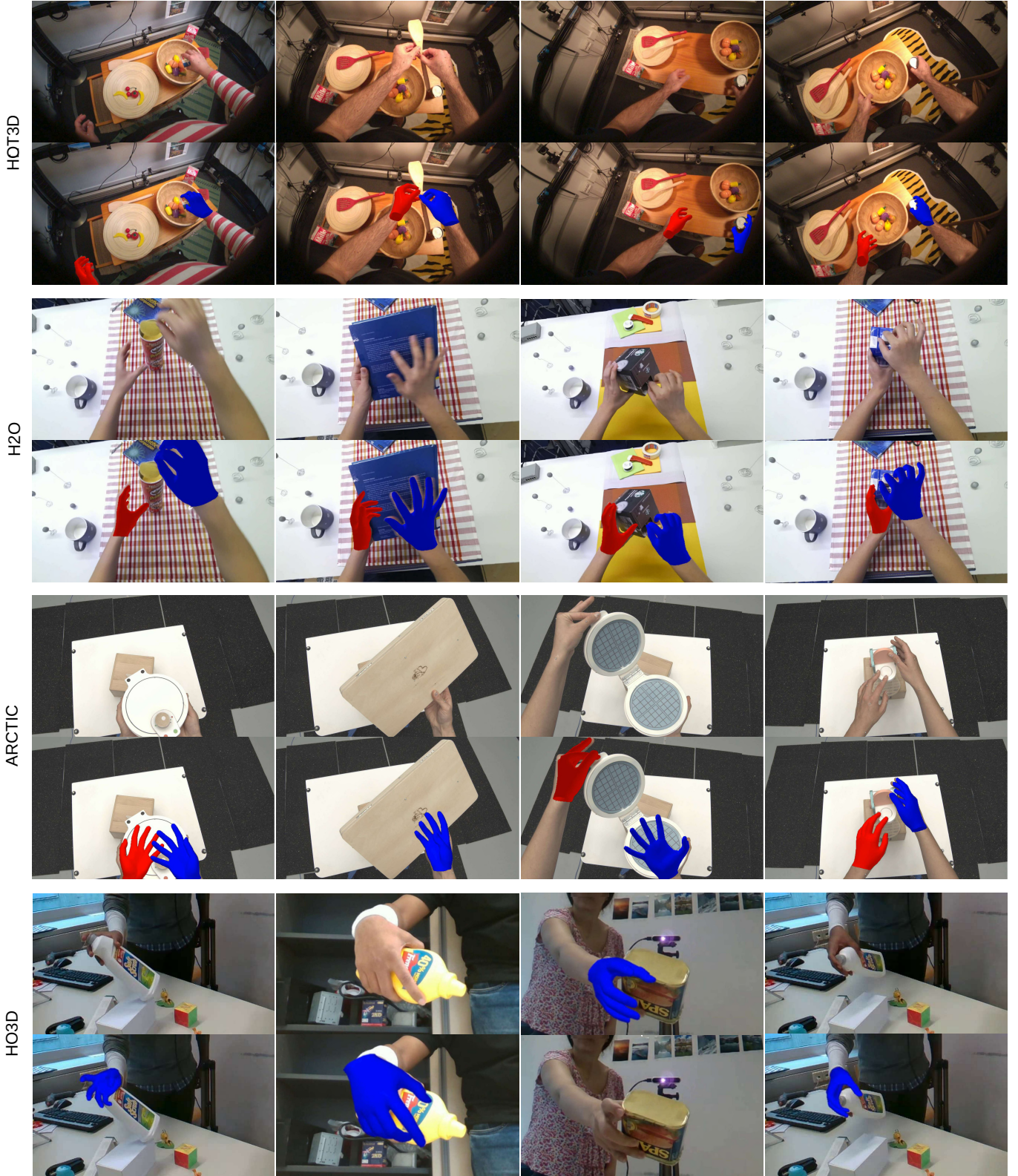}
    \vspace{-2.4em}
    \caption{\textbf{Camera-space hand mesh projections on egocentric datasets.}
    We project predicted hand meshes onto images from three camera types: HOT3D (fisheye), H2O/HO3D (perspective), and ARCTIC (distorted perspective).
    Our method maintains accurate projections under challenging conditions such as motion blur (H2O) and hand-object occlusions (HOT3D, HO3D, ARCTIC).
    }    
    \label{fig:qualitative_comparision_datasets_hand_mesh}
\end{figure*}

\clearpage
\bibliographystyle{ACM-Reference-Format}
\bibliography{citations}

\clearpage

\begin{strip}
  \begin{center}
    {\Huge \textsc{Supplementary Material}}
    \\[2em]
  \end{center}

  \smallskip
  \rule{\textwidth}{0.5pt}

  \vspace{3em}

  \includegraphics[width=\textwidth]{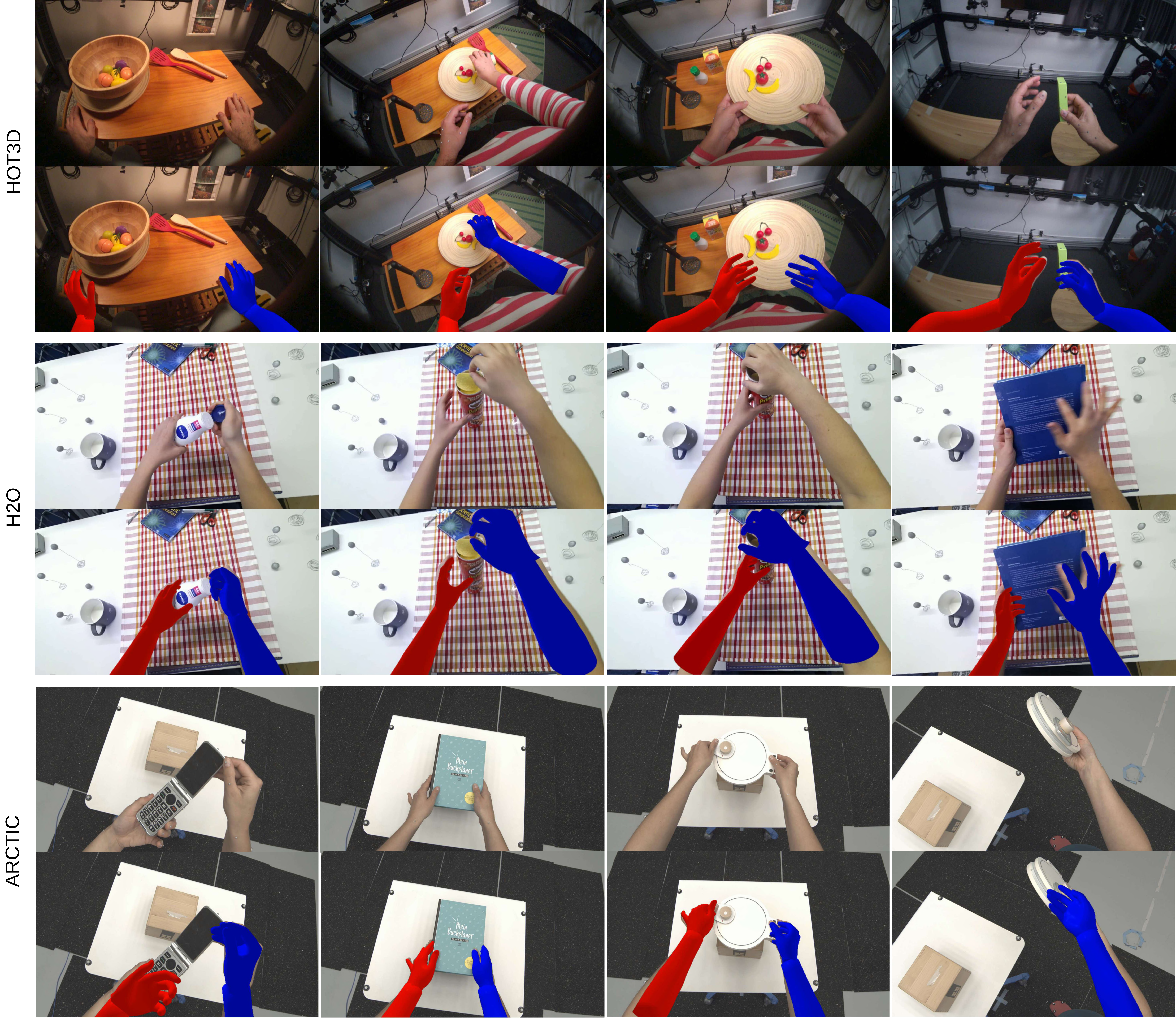}
  \vspace{-2.1em}
  \captionof{figure}{\textbf{Camera-space hand-arm mesh projections on egocentric datasets.}
  We project predicted hand and arm meshes onto images from three camera types: HOT3D (fisheye), H2O (perspective), and ARCTIC (distorted perspective).
  Our method maintains accurate projections under challenging conditions such as motion blur (H2O) and hand-object occlusions (HOT3D, ARCTIC).}
  \label{fig:qualitative_comparision_datasets_hand_arm_mesh}
  
\end{strip}

\begin{figure}[t]
    \centering
    \includegraphics[width=\linewidth]{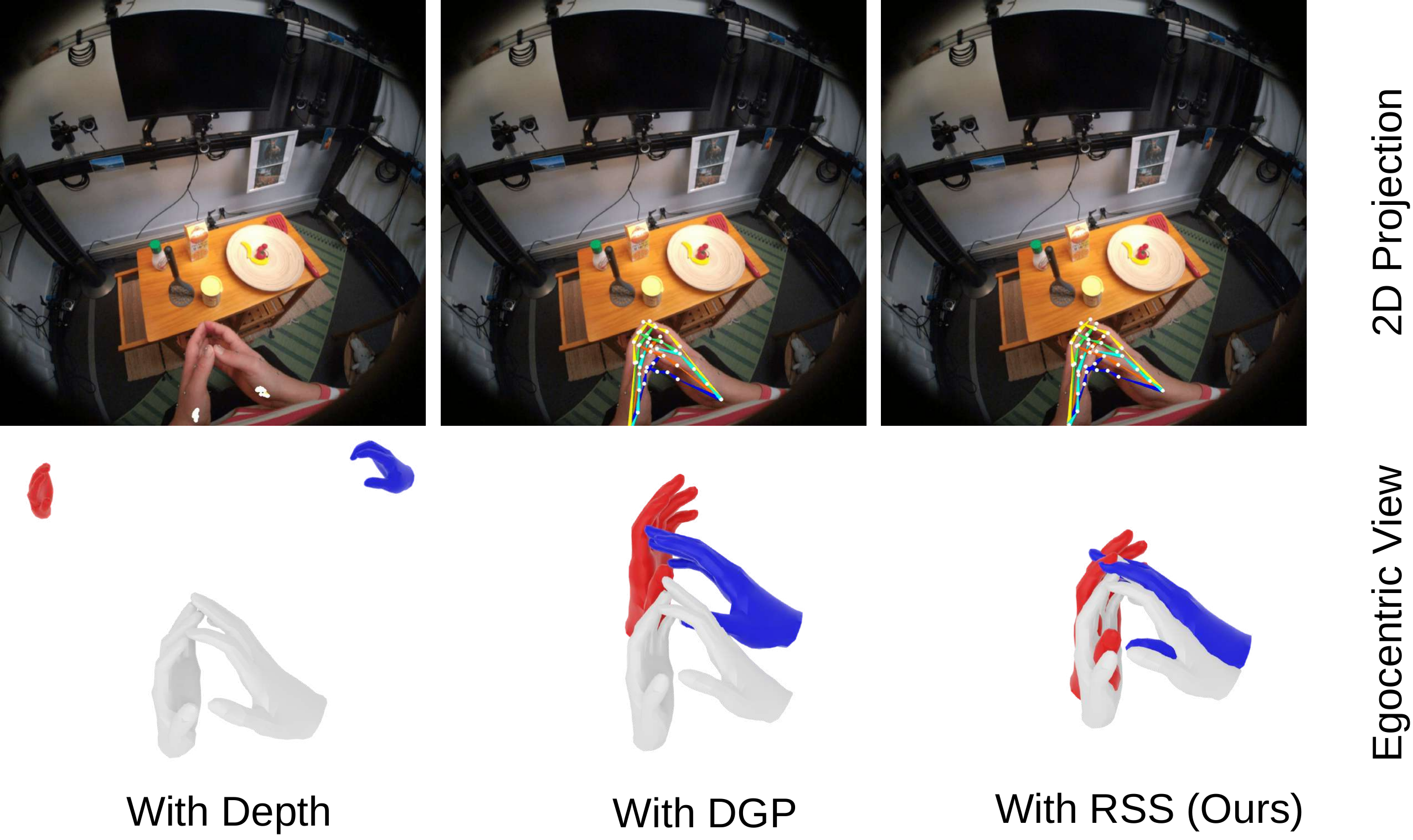}
    \vspace{-2.4em}
    \caption{\textbf{Camera-space lifting techniques.}
    Lifting via monocular depth estimators is brittle in the near field of the camera, where small depth errors lead to large camera-space misplacements.
    HandDGP’s DGP module can yield plausible 2D projections while placing the 3D hand mesh far from the ground truth.
    In contrast, our RSS produces both accurate 2D projections and camera-space mesh placements that closely match ground truth.
    }
    \label{fig:camera_space_lifting_ablation}
\end{figure}

\section{Additional Details about our Framework}

\subsection{ForeArm Representation Model}
\label{sec:FARM_repr}
The ForeArm Representation Model (FARM) is a lightweight, fully differentiable, parameterized mesh generator that maps a low-dimensional parameter vector to a watertight triangular mesh $\bigl(\mathbf V,\ \mathbb F\bigr)$ suitable for modeling individual limb segments.
Geometrically, FARM approximates the limb as a truncated cone and defines three anatomically meaningful 3D joints along its length. 
In the forearm configuration, these joints correspond to the elbow, mid-forearm, and wrist.

\subsubsection{Construction}

FARM constructs its mesh by sampling vertices on a regular angular–height lattice:

\[
  \theta_i = \frac{2\pi\,i}{n_\theta}, \quad
  z_j = \frac{j}{n_z - 1}\,h,
\]

\[
  i = 0,\dots,n_\theta - 1,\quad
  j = 0,\dots,n_z - 1,
\]
where $n_\theta, n_z \in \mathbb{N}$ are the numbers of angular and height subdivisions (e.g.\ $n_\theta=50$, $n_z=12$ in our implementation).

At each height level~$z_j$, the radius is given by
\[
  r_j \;=\; 
    r_1 
    + \bigl(r_2 - r_1\bigr)\,\frac{j}{n_z - 1}
    + \rho_j,
  \qquad
  \bm\rho = \bigl(\rho_0, \dots, \rho_{n_z-1}\bigr)^\top,
\]
so that the vector $\bm\rho$ serves as a learned radial offset profile, allowing fine sculpting of the limb’s cross‐section.

We collect the $(x,y,z)$ coordinates as 

\[
  v_{i,j}
  = 
  \begin{bmatrix}
    r_j\cos\theta_i \\[2pt]
    r_j\sin\theta_i \\[2pt]
    z_j
  \end{bmatrix}
  \in\mathbb{R}^3,
  \quad
  V = \bigl\{v_{i,j}\bigr\}_{\substack{i=0,\dots,n_\theta-1 \\ j=0,\dots,n_z-1}}.
\]

Two additional vertices \(v_{\mathrm{bottom}}\) and \(v_{\mathrm{top}}\) cap the ends, and one midpoint vertex \(v_{\mathrm{mid}}\) is placed at \(z=\tfrac12h\).  
The face set \(\mathcal F\) comprises \(2n_\theta(n_z-1)\) quadrilaterals---each divided into two triangles---and \(n_\theta\) triangular faces on each end cap.

\subsubsection{Pose}

Let \(\bm\Omega\in\mathrm{SO}(3)\) be an arbitrary rotation, which we decompose into a \emph{swirl} component \(\bm\Omega_s\) and a \emph{twist} component \(\bm\Omega_t\):
\[
    \bm\Omega
    \;=\;
    \underbrace{\bm\Omega_s}_{\text{swirl}}
    \;\underbrace{\bm\Omega_t}_{\text{twist}}.
\]
Since the truncated‐cone geometry is radially symmetric, forearm pronation (twist around the longitudinal axis) cannot be determined from the mesh. 
We therefore discard the twist component and retain only the swirl:
\[
  \bm\Omega_s = \bm\Omega\,\bm\Omega_t^{\!-1}.
\]

To apply the rigid transform to each vertex \(v_k\), we first recenter it about the midpoint
\[
  \mathbf c = \begin{bmatrix} 0 \\[3pt] 0 \\[3pt] \tfrac12\,h \end{bmatrix},
\]
then rotate and translate:
\[
  \tilde{\mathbf v}_{i,j}
  = \bm\Omega_s\,\bigl(\mathbf v_{i,j} - \mathbf c\bigr)
    + \mathbf c + \mathbf t,
\]
where \(\mathbf t\in\mathbb R^3\) is the global translation of the FARM mesh.
The same transformation is applied to the three joint centres (elbow, mid‐forearm, and wrist), making them suitable for extracting FARM parameters from existing datasets.

\subsubsection{Low‑Dimensional Shape Space}

The FARM shape parameters are highly expressive, which can make optimization---when relying solely on sparse 3D joints and segmentation masks---ill-posed. 
To address this, we impose a geometric prior by learning a PCA model over the parameter vector
\[
  \mathbf s
  \;=\;
  \begin{bmatrix}
    r_1 & r_2 & h & \rho_0 & \dots & \rho_{\,n_z-1}
  \end{bmatrix}^{\!\top}
  \;\in\;\mathbb R^{3+n_z},
\]

Specifically, we introduce a low-dimensional latent code \(\mathbf p \in \mathbb{R}^d\) (with \(d=5\)) and decode it linearly:
\[
  \mathbf s = \mathbf W\,\mathbf p + \mathbf b,
\]
where \(\mathbf W \in \mathbb{R}^{(3+n_z)\times d}\) is the PCA loading matrix and \(\mathbf b \in \mathbb{R}^{3+n_z}\) is the mean. 
Both \(\mathbf W\) and \(\mathbf b\) are learned from data and then frozen to regularize the forearm shape toward plausible geometries.

\noindent\textbf{PCA Space Training.}
The PCA loading matrix \(\mathbf W\) and mean vector \(\mathbf b\) are computed from forearm parameter vectors extracted from the AMASS motion‐capture repository~\cite{AMASS_ICCV_2019}. 
Specifically, we sample $2806$ SMPL body meshes---covering 344 unique subjects performing a wide variety of motions---and isolate the corresponding forearm parameters $s$. 
We then apply principal component analysis to this collection and retain the top \(d=5\) components, which together capture approximately $99\%$ of the variance in forearm shape, to form \(\mathbf W\) and \(\mathbf b\).

\noindent\textbf{SMPL to FARM Fitting.} For each SMPL sample, we perform the following steps:

\begin{enumerate}
  \item \textbf{Extract forearm vertex set.}  
    Select the SMPL vertices whose 3D coordinates lie between the anatomical elbow and wrist joint centers. Denote this set by
    \[
      V = \{\,v_i \mid v_i \text{ lies between elbow and wrist}\}.
    \]
  \item \textbf{Estimate initial parameters.}  
    From \(V\), extract the two boundary rings
    \[
      V_{\mathrm{elbow}}  = \{\,v_i \mid i \in I_{\mathrm{elbow}}\}, 
      \quad
      V_{\mathrm{wrist}} = \{\,v_i \mid i \in I_{\mathrm{wrist}}\},
    \]
    and let
    \[
      V_{\mathrm{ring}} = V_{\mathrm{elbow}} \,\cup\, V_{\mathrm{wrist}}.
    \]
    Perform PCA on these rings to estimate
    \[
      h,\quad r_1,\quad r_2
    \]
    where \(h\) is the forearm length, \(r_1\) the elbow radius, and \(r_2\) the wrist radius.

  \item \textbf{Instantiate FARM.}  
    Generate the FARM mesh with \((n_\theta=50,\,n_z=10)\) using the initial shape parameters \(\{r_1,r_2,h\}\) and setting \(\bm\rho=\mathbf0\).  
    Extract the elbow and wrist boundary indices \(\hat I_{\mathrm{elbow}}, \hat I_{\mathrm{wrist}}\) and form the ring set
    \[
      \hat V_k = \bigl\{\hat v_i \mid i \in \hat I_{\mathrm{elbow}}\cup \hat I_{\mathrm{wrist}}\bigr\}.
    \]
  \item \textbf{Pose optimization.}  
    Keeping the shape parameters \(\mathbf s\) fixed, optimize only the global rotation \(\bm\Omega\) and translation \(\mathbf t\) by minimizing
    \[
      \mathcal L_{\rm pose}
      = \lambda_k\,d_{\rm Chamfer}\bigl(\hat V_k,\,V_k\bigr)
      + \lambda_v\,d_{\rm Chamfer}\bigl(\hat V,\,V\bigr),
    \]
    with \(\lambda_k=100\) and \(\lambda_v=10\).  
    We employ the Adam optimizer (learning rate \(0.1\)), a ReduceLROnPlateau scheduler (factor \(0.9\), patience \(10\)), and early stopping (patience \(100\), \(\Delta_{\min}=10^{-6}\)).  
    Optimization terminates when the early-stopping criterion is met, yielding \(\bm\Omega^*\) and \(\mathbf t^*\).

\item \textbf{Shape optimization.}  
We freeze the optimal pose \(\bm\Omega^*, \mathbf t^*\) and optimize the shape parameters \(\mathbf s = [r_1,\,r_2,\,h,\,\bm\rho]^\top\) by minimizing
\[
  \begin{aligned}
    \mathcal L_{\rm shape}
    &= \alpha_k\,d_{\rm Chamfer}\bigl(\hat V_k,\,V_k\bigr)\\
    &\quad + \alpha_v\,d_{\rm Chamfer}\bigl(\hat V,\,V\bigr)\\
    &\quad + \alpha_{\rm vol}\,\Bigl|\mathrm{Vol}(r_1,r_2,h,\bm\rho) - V_{\rm mesh}\Bigr|\\
    &\quad + \alpha_{\Delta r}\,\bigl|r_2 - r_1\bigr|
     + \alpha_{1}\,\bigl|r_1 - r_1^{(0)}\bigr|
     + \alpha_{2}\,\bigl|r_2 - r_2^{(0)}\bigr|,
  \end{aligned}
\]
where the FARM volume 
$$
  \mathrm{Vol}(r_1,r_2,h,\bm\rho)
  = \frac{\pi}{3}\sum_{j=0}^{n_z-2}\Delta z\,\bigl(r_j^2 + r_j\,r_{j+1} + r_{j+1}^2\bigr),
$$
$$
  \Delta z = \frac{h}{n_z-1},
$$
is computed as the sum of frusta and the weights are set to
\(\alpha_k=10,\;\alpha_v=1,\;\alpha_{\rm vol}=1,\;\alpha_{\Delta r}=-0.1,\;\alpha_{1,2}=0.01\).
We employ the Adam optimizer with learning rates
$
  \{r_1,\,r_2,\,h\}:0.001,\quad \bm\rho:0.01,
$
together with the same ReduceLROnPlateau scheduler and early stopping as in pose fitting. 
Optimization terminates when the early-stopping criterion is met, yielding the final parameters
\(\{r_1^*,\,r_2^*,\,h^*,\,\bm\rho^*\}\).

\item \textbf{Optimized parameters.}  
After completing both pose and shape optimization, the final set of FARM parameters is
\[
  \bigl\{\,r_1^*,\;r_2^*,\;h^*,\;\bm\rho^*,\;\bm\Omega^*,\;\mathbf t^*\,\bigr\}.
\]

\end{enumerate}

\subsubsection{Discussion.}
While FARM offers a compact, differentiable forearm representation, it makes several simplifying assumptions that may limit its fidelity. 
By modeling the radius and ulna as a single rigid segment, FARM cannot capture true pronation and supination---motions of up to $\pm 90^\circ$---and discards all axial twist. 
Its PCA shape prior is learned from adult SMPL meshes (AMASS), which may not generalize to children, individuals with atypical anatomy, or amputees. 
To mitigate these limitations, one can: (1) place off-axis markers on the forearm surface to break axial symmetry and make twist observable; 
or (2) incorporate inexpensive inertial measurement units (IMUs) or EMG straps on the limb to directly measure axial rotation.

\subsection{Crop Intrinsics Token}
\label{sec:CIT}

To encode the geometric context of each cropped image patch relative to its camera, we build upon the \emph{Keypoint Encodings} (KPE) of Prakash\,\etal~\cite{prakash20243d} and extend them with a set of camera-independent distribution parameters.  
Concretely, for each crop we compute the normalized principal-point offset, the crop ratios, and the half-field-of-view angles, and then concatenate these quantities with the original KPE vectors to form our Crop Intrinsics Tokens (CIT).
By conditioning the framework on CIT, we enable training on heterogeneous datasets---from wide-FOV egocentric videos to narrow-FOV third-person captures---while remaining robust to camera-specific variations, thereby improving cross-domain generalization..  

\subsubsection{Crop Intrinsics \& Distortion Correction}

Let the camera intrinsics be
\[
K = 
\begin{pmatrix}
f_x & 0   & c_x\\
0   & f_y & c_y\\
0   & 0   & 1
\end{pmatrix},
\]
where \(f_x, f_y\) are the focal lengths (in pixels) and \((c_x,c_y)\) is the principal point.  Let 
\[
d = (d_1,\dots,d_m)
\]
denote the distortion parameters of the chosen lens model (e.g., Rational Polynomial for ARCTIC, Kannala–Brandt for Reinterhand, FisheyeRadTanThinPrism for HOT3D).  

A distorted image point  
\[
P' = (u',v')^\top \;\in\;\mathcal I,
\]
where
\[
\mathcal I = \{(u,v)\in\mathbb R^2 \mid 0 \le u < W,\;0 \le v < H\},
\]
is mapped to its undistorted counterpart  
\[
P = (u,v)^\top \;\in\;\mathcal I
\]
by the distortion‐correction function  
\[
\phi_{d} : \mathcal I \;\longrightarrow\; \mathcal I,
\qquad
\phi_d(u',v') = 
\begin{pmatrix}
u\\
v
\end{pmatrix}.
\]
Here, \(\phi_d\) implements the inverse of the selected distortion model (e.g.\ Rational Polynomial, Kannala–Brandt, or FisheyeRadTanThinPrism), ensuring that all subsequent projection and cropping operations use geometrically accurate, undistorted coordinates within the image domain \(\mathcal I\).

\subsubsection{Spatial Context via Crop Geometry}
Let the hand bounding box in image \(\mathcal I\) be given by the pixel coordinates
\[
(x'_1,\,y'_1,\,x'_2,\,y'_2).
\]
Its width and height are then
\[
w' = x'_2 - x'_1,
\qquad
h' = y'_2 - y'_1.
\]
The center of this box in the distorted image is
\[
(x'_c,\,y'_c)
\;=\;
\Bigl(\tfrac{x'_1 + x'_2}{2},\;\tfrac{y'_1 + y'_2}{2}\Bigr).
\]
Applying the inverse‐distortion mapping \(\phi_{d}:\mathcal{I}\to\mathcal{I}\) to the center yields the undistorted crop midpoint:
\[
(u_c,\,v_c)
=\phi_{d}\bigl(x'_c,\,y'_c\bigr).
\]
Similarly, the undistorted coordinates of the four crop corners are
\[
\begin{aligned}
(u_{11}, v_{11}) &= \phi_{d}(x'_1, y'_1), &
(u_{12}, v_{12}) &= \phi_{d}(x'_1, y'_2),\\
(u_{21}, v_{21}) &= \phi_{d}(x'_2, y'_1), &
(u_{22}, v_{22}) &= \phi_{d}(x'_2, y'_2).
\end{aligned}
\]
Hence, the undistorted width \(w\) and height \(h\) of the hand crop are
\[
w = u_{22} - u_{11},
\qquad
h = v_{22} - v_{11}.
\]

\subsubsection{CIT Formulation}

For any undistorted image coordinate \((u,v)\in\mathcal{I}\), the local viewing direction is defined as in \citet{prakash20243d}:
\[
\boldsymbol{\theta}(u,v)
\;=\;
\begin{pmatrix}
\theta_u\\[4pt]
\theta_v
\end{pmatrix}
=
\begin{pmatrix}
\arctan\!\bigl(\tfrac{u - c_x}{f_x}\bigr)\\[6pt]
\arctan\!\bigl(\tfrac{v - c_y}{f_y}\bigr)
\end{pmatrix}.
\]
We evaluate this mapping at the undistorted crop midpoint \((u_c, v_c)\):
\[
\boldsymbol\theta_c \;=\; \boldsymbol\theta(u_c, v_c),
\]
and at each undistorted corner \((u_{ij}, v_{ij})\) for \(i,j \in \{1,2\}\):
\[
\boldsymbol\theta_{ij} \;=\; \boldsymbol\theta(u_{ij}, v_{ij}).
\]
Together, \(\boldsymbol\theta_c\) and the set \(\{\boldsymbol\theta_{ij}\}\) specify the viewing directions at the crop’s center and its four corners.

Next, we compute the six scale-normalized crop intrinsics:
\begin{align}
p_x &= \frac{c_x - u_c}{w}, 
& 
p_y &= \frac{c_y - v_c}{h}, 
\label{eq:pxpy}\\
r_w &= \frac{w}{W}, 
& 
r_h &= \frac{h}{H}, 
\label{eq:rw}\\
\alpha_x &= \arctan\!\Bigl(\frac{W}{2\,f_x}\Bigr),
& 
\alpha_y &= \arctan\!\Bigl(\frac{H}{2\,f_y}\Bigr).
\label{eq:fov}
\end{align}

Finally, we concatenate the five local-ray angles
\(\boldsymbol\theta_c,\;\boldsymbol\theta_{11},\;\boldsymbol\theta_{12},\;\boldsymbol\theta_{21},\;\boldsymbol\theta_{22}\)
(each a \(\mathbb{R}^2\) vector) with these six intrinsics values into a single vector:
\[
\mathrm{CI} = 
\begin{bmatrix}
\boldsymbol\theta_c\\
\boldsymbol\theta_{11}\\
\boldsymbol\theta_{12}\\
\boldsymbol\theta_{21}\\
\boldsymbol\theta_{22}\\[4pt]
p_x\\
p_y\\
\log r_w\\
\log r_h\\
\alpha_x\\
\alpha_y
\end{bmatrix}
\in \mathbb{R}^{16}.
\]

We then apply the standard sinusoidal positional encoding (as in \citet{vaswani2017attention}) to \(\mathbf{CI}\), yielding the \emph{Crop Intrinsics Token}
\[
\mathrm{CIT} = \mathrm{PE}(\mathbf{u}) \;\in\; \mathbb{R}^{128}.
\]

\subsubsection{Semantic Interpretation of CIT Components}
\begin{itemize}
  \item[\(\boldsymbol\theta_c,\{\boldsymbol\theta_{ij}\}\):]  
    \textbf{Local viewing directions.}  
    Each \(\boldsymbol\theta\in\mathbb R^2\) encodes the angular offset  
    of a ray from the optical axis at a specific point in the crop (center  
    or corner). 
    By supplying these five directions, the network knows the  
    precise local geometry of the patch---how objects tilt or foreshorten---  
    which is essential for accurate 3-D pose recovery.

\item[{\(p_{x,y}\):}]

\textbf{Principal-point offset.}  
Recall that the camera’s principal point \((c_x,c_y)\) is the projection of the \emph{optical axis} onto the image plane. The principal-point offset (Eqn.~\eqref{eq:pxpy}) locates the centre of projection within the patch in \([0,1]^2\).  

Why is this important?  Suppose a hand keypoint moves by \(\Delta u\) pixels to the right in the cropped image.  
Two things could cause this:
\begin{itemize}
    \item The hand really moved to the right in 3-D space.
    \item The camera crop shifted to the left (i.e.\ \(x_1\) increased).
\end{itemize}

Without \(p_x\), the network has no way to tell these apart and must implicitly learn a mapping from appearance alone, which couples object motion and crop translation.  By supplying \(p_x\):

\begin{itemize}
    \item A change in \(x_1\) (crop shift) changes \(p_x\) but not the underlying feature activations for the hand—so the network can \emph{subtract out} cropping effects.
    \item A genuine hand motion changes the relative position of pixels \emph{and} the geometric residual after compensating for \(p_x\), so the network correctly attributes that to 3-D movement.
\end{itemize}

This explicit disambiguation dramatically reduces bias when training on datasets with heterogeneous cropping strategies: the network no longer “confuses” camera recentering for object translation.

\item[{\(\log r_{w,h}\):}]
\textbf{Scale ratios.}  
The logarithms of the crop’s width and height relative to the full image inform the network about how ``zoomed in'' a given patch is after resizing it to the fixed input resolution (e.g., \(224\times224\)). 
Concretely:
\begin{itemize}
\item If the patch occupies a large fraction of the original image (\(r_w \approx 1\)), resizing this large patch down to the fixed input size effectively reduces the region's apparent size.
  
\item Conversely, if the patch is very small relative to the original image (\(r_w \ll 1\)), resizing that small patch to the fixed input significantly magnifies the region.
\end{itemize}

Expressing these ratios in log-space enables the network to linearly and smoothly interpolate across a wide range of zoom levels, facilitating robust generalization to diverse cropping strategies.

\item[{\(\alpha_{x,y}\):}]
\textbf{Half–field–of–view angles.}  
  These angles define the camera’s total angular aperture in the horizontal
  and vertical directions, and can be interpreted as the sensor’s
  angular resolution per pixel
  For instance, a horizontal displacement of \(\Delta u\) pixels on the
  image plane corresponds to an angular change of \(\Delta u/f_x\) radians.
  Supplying \((\alpha_x,\alpha_y)\) therefore allows the network to
  convert image‑space displacements into real‑world directions in a
  device‑independent manner.  

\end{itemize}

Hence, each component of the CIT plays a complementary role in achieving our objective of camera-model general, cross-camera 3D hand pose estimation.

\begin{figure}[h]
  \centering
  \includegraphics[width=0.8\linewidth]{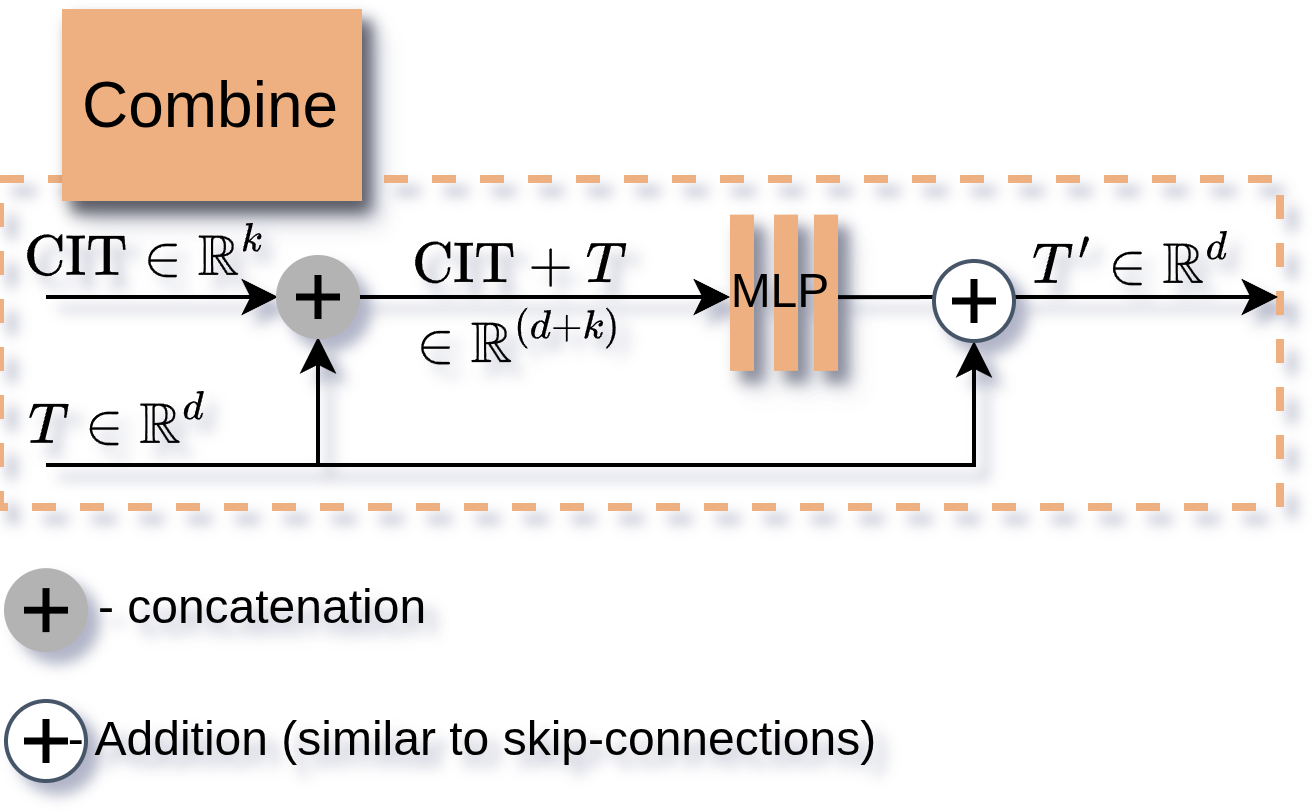} 
  \vspace{-10pt}
  \caption{
  \textbf{Fusing CIT and Crop Tokens.}
For each crop (hand/arm), its CIT is broadcast to all patch tokens for that crop and fused in the Combine block via feature concatenation followed by a learnable projection, with a residual addition of the original token embedding.
This injects crop-specific geometric context into every patch feature while allowing the model to fall back to the original mapping when the conditioning is unnecessary.
}
  \label{fig:CaDBlock}
  \vspace{-1em}
\end{figure}

\subsection{Ray Space Solver (RSS)}
\label{sec:ray_depth_solve}

The Ray Space Solver (RSS) takes the 3D joints positions $\bm J_i$, their corresponding estimated 2D image keypoints ($u_i,v_i$), and the associated 2D confidence weights $w_i$, to compute the camera-space translation $\mathbf t$ by enforcing that each joint lies along its corresponding viewing ray:
\begin{equation}
    \mathbf J_i + \mathbf t \;=\; \lambda_i \,\mathbf d_i,
    \qquad i = 1,\dots,M,
    \label{eq:point-on-ray}
\end{equation}
is weighted by $w_i$, where $M$ is the number of joints and $\bm d_i$ is the unit‐direction vector of the ray passing through the 2D keypoint ($u_i,v_i$).

To compute each ray direction  $\bm d_i$, we first back‐project the 2D keypoint ($u_i,v_i$) into a normalized camera coordinate frame:

$$
  \bar u_i \;=\;\frac{u_i-c_x}{f_x},\qquad
  \bar v_i \;=\;\frac{v_i-c_y}{f_y},\qquad
    \rho_i      = \sqrt{\bar u_i^{2}+\bar v_i^{2}},
$$

where $f_x, f_y$ are the focal lengths and ($c_x, c_y$) is the principal point. 
For a pinhole model,
$$
\tilde{\mathbf d}_i = (\bar u_i,\bar v_i,1)^\top,\quad
\mathbf d_i=\tilde{\mathbf d}_i/\|\tilde{\mathbf d}_i\|.
$$
For an equidistant fisheye (example),
$$
\mathbf d_i=\Big(\tfrac{\bar u_i}{\rho_i}\sin \rho_i,\ \tfrac{\bar v_i}{\rho_i}\sin \rho_i,\ \cos \rho_i\Big)^\top.
$$
(Other calibrated models, e.g., rational polynomial or Kannala–Brandt, provide a similar unprojection function; in all cases we finally L2-normalize $\mathbf d_i$.)

A weighted least-squares fit of \eqref{eq:point-on-ray} can be formulated as
\begin{equation}
    \min_{\mathbf t,{\lambda}} ;\sum_{i=1}^M w_i ,\big\| (\mathbf t+\mathbf J_i) - \lambda_i \mathbf d_i \big\|^2 
\label{eq:rearrange-point-on-ray}
\end{equation}
for each joint $i=1,\dots,M$. For fixed $\mathbf t$, the optimal depth obtained in closed form:
\begin{equation}
\label{eq:lambda_star}
\lambda_i^\star \;=\; \mathbf d_i^\top(\mathbf t+\mathbf J_i).
\end{equation}
Substituting \eqref{eq:lambda_star} into \eqref{eq:rearrange-point-on-ray} yields the \emph{point-to-ray least-squares} objective
\begin{equation}
\label{eq:rdls}
\min_{\mathbf t}\; E(\mathbf t)=\sum_{i=1}^M w_i\,\big\|\mathbf r_i\big\|^2,
\qquad
\mathbf r_i \;=\; \underbrace{\big(I-\mathbf d_i\mathbf d_i^\top\big)}_{\mathbf\Pi_i}\,(\mathbf t+\mathbf J_i).
\end{equation}
Here $\mathbf\Pi_i$ is the orthogonal projector onto the plane perpendicular to $\mathbf d_i$ and maps any vector to its component perpendicular to $\mathbf d_i$. 
In particular,
$$
\mathbf\Pi_i\mathbf d_i \;=\; \big(I-\mathbf d_i\mathbf d_i^\top\big)\mathbf d_i
= \mathbf d_i - \mathbf d_i(\mathbf d_i^\top\mathbf d_i)
= \mathbf d_i - \mathbf d_i
= \mathbf 0,
$$
so $\mathbf\Pi_i$ \emph{annihilates} all along-ray (depth) components. 
For any $\mathbf v\in\mathbb R^3$,
\begin{align}
\mathbf v_{\parallel} &= (\mathbf v^\top\mathbf d_i)\,\mathbf d_i 
&&\text{(``depth'' component)},\nonumber\\
\mathbf v_{\perp}     &= \mathbf v - \mathbf v_{\parallel} = (I-\mathbf d_i\mathbf d_i^\top)\mathbf v = \mathbf\Pi_i\mathbf v 
&&\text{(``sideways'' component)}.\nonumber
\end{align}
Hence $\|\mathbf\Pi_i\mathbf v\|=\|\mathbf v\|\sin\theta=\|\mathbf v\times \mathbf d_i\|$, where $\theta$ is the angle between $\mathbf v$ and $\mathbf d_i$.

In Eqn.~\ref{eq:rdls}, $\mathbf r_i=\mathbf\Pi_i(\mathbf t+\mathbf J_i)$ is the sideways residual component: it is exactly the shortest vector from the translated point $\mathbf t+\mathbf J_i$ to the ray through $\mathbf d_i$.

Differentiating \eqref{eq:rdls} w.r.t.\ $\mathbf t$ gives the $3\times 3$ weighted equations
\begin{equation}
\label{eq:normal_eq}
\Big(\sum_{i=1}^M w_i\,\mathbf\Pi_i\Big)\mathbf t \;=\; -\sum_{i=1}^M w_i\,\mathbf\Pi_i\,\mathbf J_i,
\qquad\text{i.e.,}\qquad
M\,\mathbf t \;=\; -\,\mathbf m,
\end{equation}
with
\[
M=\sum_{i=1}^M w_i \mathbf\Pi_i \in \mathbb R^{3\times 3},
\qquad
\mathbf m=\sum_{i=1}^M w_i \mathbf\Pi_i \mathbf J_i \in \mathbb R^3.
\]
We solve \eqref{eq:normal_eq} using a tiny Tikhonov term for stability:
\begin{equation}
\label{eq:solve_t}
\widehat{\mathbf t}
\;=\;
-\,(M+\varepsilon I)^{-1}\mathbf m,
\qquad \varepsilon\ll 1.
\end{equation}

After $\widehat{\mathbf t}$ is found, each joint’s camera-space position are
$$
\mathbf p_i=\widehat{\mathbf t}+\mathbf J_i,
$$

\paragraph{Notes on robustness.}
In practice we check the condition number $\kappa(M)$; if $\kappa(M)\ge 10^6$ we damp the weights ($w_i\!\leftarrow\!0.5\,w_i$) and re-solve.
For more details about the formulation, we refer the readers to Tikhonov regularization~\cite{hoerl1970ridge}.

\subsubsection{Kalman Filter}
\label{supp:sec:kalman}
We use a 3D constant-velocity Kalman filter on the predicted camera-space translations. 
The process and measurement noise variances \((q_{\text{pos}}, q_{\text{vel}}, r_{\text{meas}})\) are tuned offline on H2O, HOT3D and ARCTIC datasets.
We select the filter hyperparameters by minimizing a simple objective that trades off (i) accuracy w.r.t.\ ground truth during visible frames and (ii) temporal smoothness:
\begin{equation}
\mathcal{L} \;=\; \lambda\,\mathcal{L}_{\text{fid}} \;+\; (1-\lambda)\,\mathcal{L}_{\text{smooth}},
\end{equation}
where
\begin{equation}
\mathcal{L}_{\text{fid}}
\;=\;
\left\lVert 
\tilde{\mathbf{p}}_{t}-\mathbf{p}^{\mathrm{gt}}_{t}
\right\rVert_2^2
\end{equation}
measures fidelity to the ground-truth translation on visible frames, and
\begin{equation}
\mathcal{L}_{\text{smooth}}
\;=\;
\left\lVert \tilde{\mathbf{p}}_{t+1}-2\tilde{\mathbf{p}}_{t}+\tilde{\mathbf{p}}_{t-1}\right\rVert_2^2
\end{equation}
penalizes second-order temporal differences (acceleration), encouraging smooth motion. We fix $\lambda$ (e.g., $\lambda=0.7$).

\section{Experimental Evaluation}

\subsection{Dataset Details}
\label{supp:subsec:dataset_splits}
\noindent
\emph{ARCTIC}~\cite{fan2023arctic} contains 393 sequences of hand–object interactions involving 11 articulated objects and 10 subjects, recorded from eight static and one egocentric fisheye view. 
We use both exocentric and egocentric frames from the official training split (1.7M images) for training, and evaluate exclusively on the egocentric subset (23K images) from the official validation split.
Since the official test set is not publicly available and does not support camera-space joint-error evaluation, we report results on the validation set as a proxy for testing.

\noindent
\emph{H2O (Two Hands and Objects)}~\cite{Kwon_2021_ICCV} provides synchronized multi-view RGB-D sequences of bimanual hand–object interactions with 3D hand and object poses, camera parameters, and meshes. 
We use both exocentric and egocentric frames from the official training split (167K images)  for training, and evaluate only on the egocentric frames from the official test split (23K images).

\noindent
\emph{HOT3D}~\cite{banerjee2024introducing} offers over 833 minutes of egocentric video from Project Aria glasses~\cite{engel2023projectarianewtool}, featuring 19 participants interacting with 33 rigid objects. We use only the monocular RGB stream and divide it into 354K training and 110K test frames (see Sup.~Sec.~\ref{supp:subsec:HOT3D_Split} for split details).

\noindent
\emph{HO3D}~\cite{hampali2020honnotate} consists of hand–object interaction sequences with severe occlusions caused by object manipulation. 
We use the V2 version of the dataset and follow the official training (66K images) and test (11K images) splits.

\noindent
For \emph{Re:InterHand} and \emph{HandCO}, we use the entire datasets for training.

\subsection{HOT3D Split}
\label{supp:subsec:HOT3D_Split}
Although HOT3D provides an official 20\% test split, we ignore it due to the lack of ground-truth annotations for testing the CS-MJE and no official server to test this metric (to the best of our knowledge); instead, we take the remaining 80\% of the data and split it into 60\% for training and 20\% for validation. 
The full list of image splits will be released upon publication.

\subsection{FARM Generation for Datasets}
\label{supp:subsec:FARM_GEN}
\noindent \textbf{H2O}. We first train a dedicated 2D arm-pose network and run it on every camera view of the H2O recordings to obtain per-view arm 2D keypoints. 
These detections are triangulated across views to recover 3D arm joints. 
In parallel, we generate multi-view arm segmentation masks. 
The resulting 3D keypoints and silhouettes are then fed to our optimisation stage, which refines arm pose and shape and returns the corresponding FARM parameters.

\noindent \textbf{ARCTIC}. For ARCTIC we directly optimise the arm vertices of an SMPL-X body model using the available 3D supervision, and convert the fitted mesh to the required FARM parameter set.

\subsection{Evaluation Metrics}
\label{supp:subsec:Evaluation_Metrics}
We report the following metrics:

\begin{itemize}[leftmargin=1em]
\item \textbf{Camera-space Mean Joint Error (CS-MJE).} 
Mean Euclidean distance (mm) between predicted and ground-truth 3D joints in the camera-space~\cite{CMR,handdgp2024}, capturing errors in hand pose, translation, scale, and rotation.
\item \textbf{Root-relative Mean Joint Error (RS-MJE).}
Mean Euclidean distance (mm) after subtracting the root joint~\cite{grauman2024ego,chen2024pcie_egohandpose}, capturing pose, scale, and rotation errors while ignoring translation.
\item \textbf{Procrustes-aligned Mean Joint Error (PS-MJE).}
Mean Euclidean distance (mm) after rigid Procrustes alignment (removing scale, rotation, and translation), following~\cite{HaMeR,MeshTransformer}.
\item \textbf{Acceleration error (ACC).}
ACC measures temporal stability \cite{kanazawa2019learning,Vibe}. We report CS-ACC in camera space (translation jitter) and RS-ACC in root-relative space (hand jitter), both in $m/s^2$.
\end{itemize}

\section{Additional Experiments}

\subsection{Results}

\begin{table}[h]
\centering
\caption{
Comparison between the released HandDGP model weights and our reimplementation ($\dagger$) on HOT3D (rectified).
Since HandDGP does not provide training code, $\dagger$ is trained via our best-effort reproduction under the same specification as HandDGP. 
Both variants are evaluated with an identical preprocessing (same rectification, cropping, resolution).
Similar results on HOT3D for the released HandDGP model weights are also reported in \citet{HaWoR}.
We report CS-MJE (mm), PS-MJE (mm), and CS-ACC (m/s$^2$); lower is better for all metrics.
}
\vspace{-0.8em}
\small
\setlength{\tabcolsep}{7pt}
\begin{tabular}{@{} l
                S[table-format=3.2]
                S[table-format=2.2]
                S[table-format=2.2] @{}}
\toprule
Method &  {CS-MJE $\downarrow$} & {PS-MJE $\downarrow$} & {CS-ACC $\downarrow$} \\
\midrule
HandDGP (released weights)       & 109.3 & 16.3 & 21.9 \\
HandDGP$\dagger$ (reimplemented) &  61.3 &  8.6 & 20.4 \\
\bottomrule
\end{tabular}
\label{tab:hand_dgp_eval}
\end{table}

\noindent \textbf{In-the-Wild.}
We define \emph{in-the-wild} video sequences as videos for which no calibrated camera model is available.
In these cases, we estimate camera intrinsics using AnyCalib~\cite{tirado2025anycalib}.
While this strategy works reasonably well for pinhole optics---as illustrated in Fig.~\ref{fig:sota_HO3D_and_in_the_wild}(b) and in the supplementary video---our experiments in Table~\ref{tab:camera_space_lifting} show that accuracy degrades when applying the same procedure to fisheye cameras.
Moreover, because our method is trained exclusively on calibrated 3D datasets recorded in controlled laboratory environments, it does not fully generalize to out-of-domain data.

\begin{figure*}[t]
  \centering
  \includegraphics[width=\linewidth]{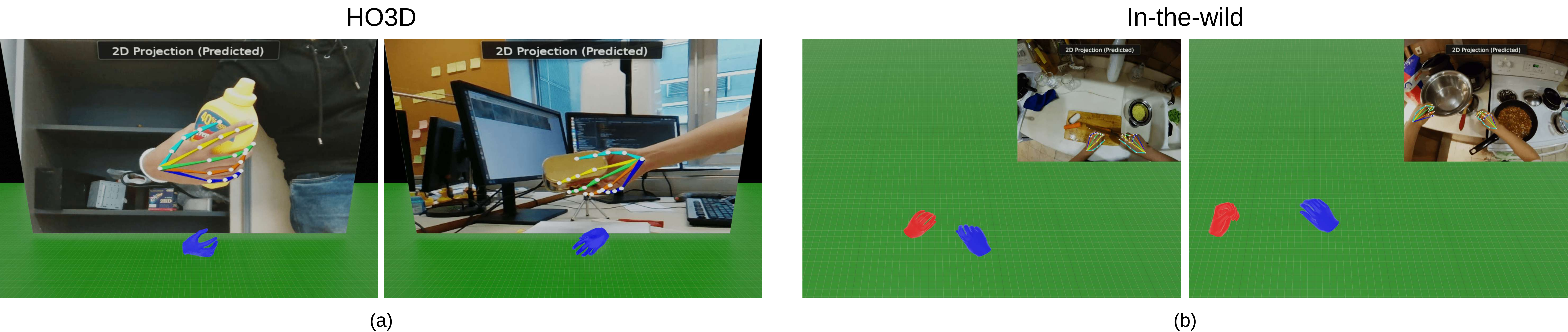} 
  \vspace{-1.8em}
  \caption{%
\textbf{Qualitative results on HO3D and in-the-wild data.}
Our approach produces accurate hand pose estimates even under hand--object occlusions on HO3D (a), and it generalizes to in-the-wild videos despite not being explicitly trained on those data distributions (b).%
}
\label{fig:sota_HO3D_and_in_the_wild}
\vspace{-1em}
\end{figure*}

\begin{figure}[h]
    \centering
    \includegraphics[width=\linewidth]{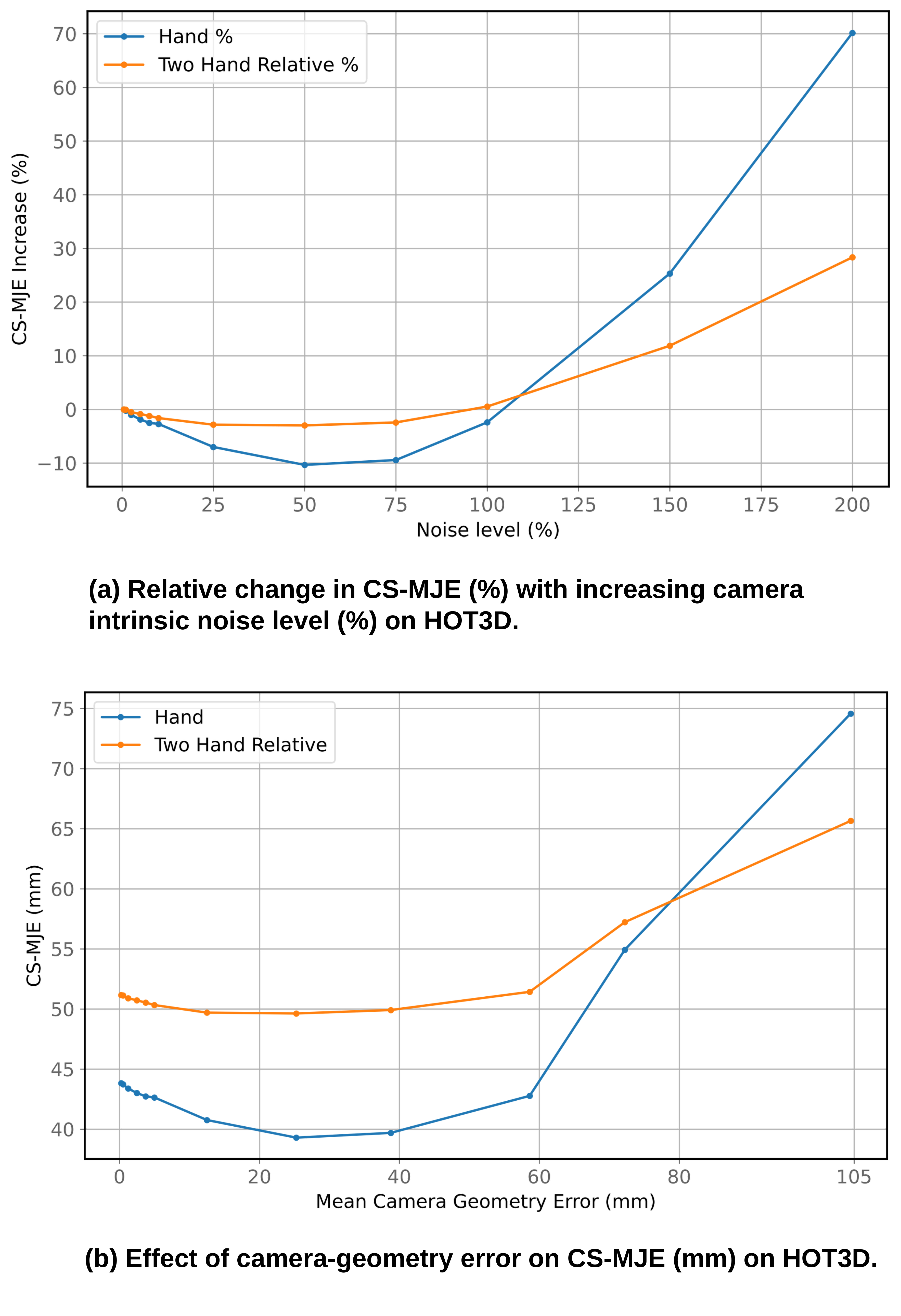}
    \vspace{-2.4em}
    \caption{\textbf{Robustness to calibration mismatch on HOT3D.}
As camera-intrinsic perturbation increases, CS-MJE remains stable and even improves slightly under moderate mismatch, despite increasing camera-geometry error; performance degrades clearly only under large mismatches, indicating robustness to moderate calibration error.
    }
    \label{fig:intrinsics_robustness}
\end{figure}

\noindent \textbf{Runtime Comparison.}
The initial hand-arm detection stage of our pipeline takes around $40$~ms to run, followed by the model forward pass at $24.2$~ms, and the RSS lifting module at $3.1$~ms. 
Tab.~\ref{tab:hot3d_runtime} reports compute time (ms) for the model forward pass and the lifting step across methods, measured with a batch size of two (both hands) on an RTX~3090.
We report the mean runtime with the standard deviation spread $\sigma$ in parentheses, where $\sigma$ denotes three standard deviations and serves as a measure of runtime jitter.
Feed-forward approaches, such as HandDGP and our \textsc{EgoForce}, exhibit consistently low lifting cost ($2.8$--$3.1$~ms), whereas optimization-based methods incur substantially higher lifting overhead, most notably HandOccNet ($220.9$~ms with large jitter), reflecting the iteration- and conditioning-dependent nature of per-frame refinement.

\begin{table}[t]
\caption{\textbf{Runtime performance metrics on the HOT3D dataset.}
We report mean compute time (ms) with the corresponding $\sigma$ deviation in parentheses.
``FF'' denotes a feed-forward model, and ``Opti'' denotes iterative optimization-based 3D-to-2D lifting.
}
\centering
\small
\setlength{\tabcolsep}{10pt}
\renewcommand{\arraystretch}{1.15}
\begin{tabular}{lcc c}
\toprule
\multirow{2}{*}{Method} & \multicolumn{2}{c}{Compute Time (ms)} & \multirow{2}{*}{Type} \\
\cmidrule(lr){2-3}
 & Model ($\sigma$) & Lifting ($\sigma$) & \\
\midrule
MobRecon   & 14.0 (8.8)  & 18.1 (12.2)  & FF + Opti\\
HandOCCNET & 15.3 (2.1)  & 220.9 (313.8)& FF + Opti\\
\midrule
HandDGP    & 17.5 (3.3)  & 2.8 (15.8)  & FF\\
OURS       & 24.2 (2.4)  & 3.1 (0.5)   & FF\\
\bottomrule
\end{tabular}
\label{tab:hot3d_runtime}
\end{table}

\noindent
\textbf{Comparison to Additional SOTAs.}
\label{subsec:additional_SOTAs}
UmeTrack~\cite{han2022umetrack} and HaWoR~\cite{HaWoR} did not release training code, providing only pretrained models and evaluation pipelines; our comparisons are based on their released inference and evaluation setups.

\begin{figure*}[h]
    \centering
    \includegraphics[width=\linewidth]{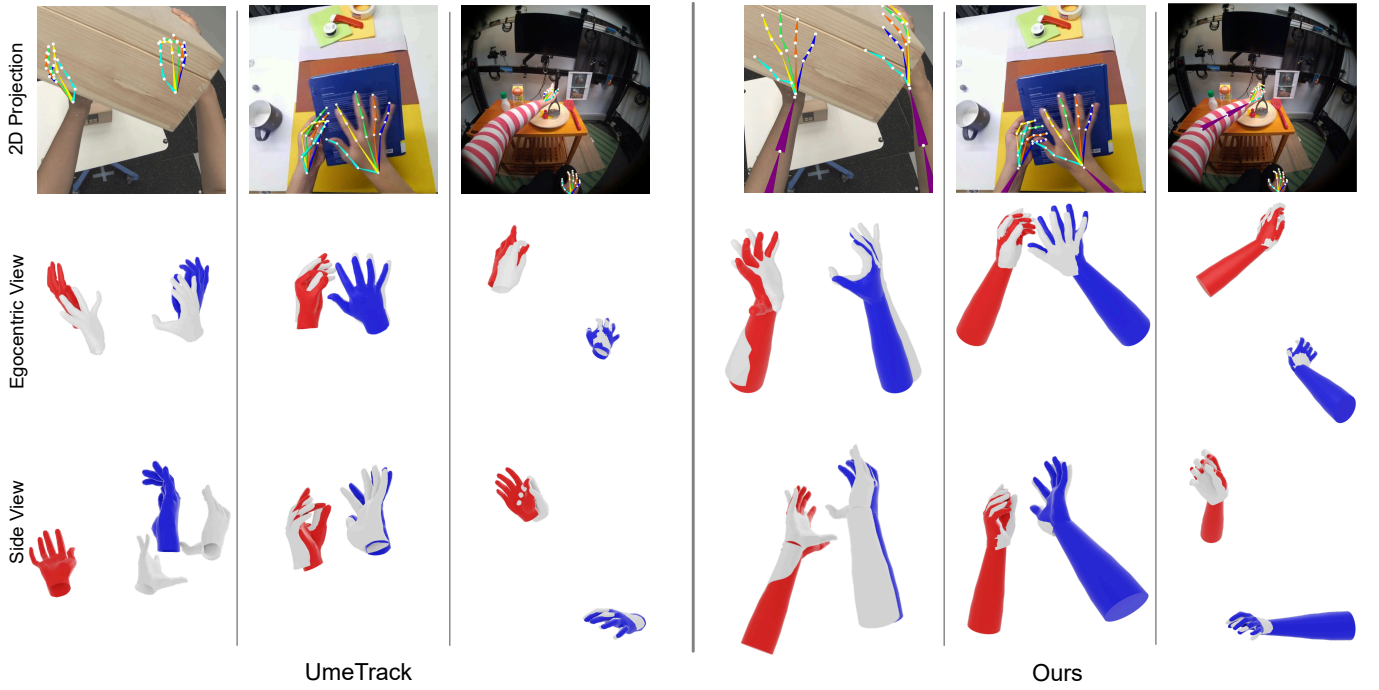}
    \vspace{-1.4em}
    \caption{\textbf{Qualitative camera-space results on egocentric datasets.}
    We compare our method against UmeTrack~\cite{han2022umetrack} on three datasets with widely different camera intrinsics.
    Predicted left and right limb meshes are shown in \textcolor{sigred}{red} and \textcolor{sigblue}{blue}, respectively, with ground truth highlighted in gray.
    }
    \label{fig:qualitative_umetrack_comparision}
    \vspace{-1em}
\end{figure*}

\paragraph{UmeTrack.} 
Table~\ref{tab:UMETRACK_results} compares EgoForce with UmeTrack under two crop settings.
When UmeTrack is evaluated with a perspective crop built from ground-truth 3D keypoints, this should be regarded as a best-case setting, since such crop information is unavailable in real deployment.
Even under this favorable protocol, EgoForce achieves lower PS-MJE on all three datasets and also lower CS-MJE on ARCTIC, showing that its gains are not limited to global translation recovery but also extend to articulated hand reconstruction.
The more realistic comparison is therefore the setting (UmeTrack*) in which the perspective crop is built from 2D hand bounding boxes, which are available at inference time either from dataset annotations or from a hand detector.
Under this realistic protocol, EgoForce is clearly superior, reducing CS-MJE by $68.1$\% on ARCTIC, $83.2$\% on HOT3D, and $79.9$\% on H2O, while also improving PS-MJE by $67.3$\%, $77.2$\%, and $74.1$\%, respectively.

Moreover, our hand-scale analysis shows that UmeTrack in single-view evaluation exhibits $6$~mm frame-wise scale variability (standard deviation across continuous frame sequences) on its own dataset and $18$~mm on HOT3D, whereas EgoForce shows only $4$~mm on HOT3D.
This gap is expected. 
UmeTrack was introduced primarily as a multi-view VR hand-tracking system, and its formulation explicitly notes that recovering hand scale for an unknown skeleton requires multi-view features, since single-view input is inherently affected by scale ambiguity.
In contrast, EgoForce is explicitly designed for monocular camera-space reconstruction: forearm cues provide metric information to reduce monocular depth--scale ambiguity, and ray-space lifting preserves calibrated camera geometry across different optics.
This further indicates that EgoForce is not only more accurate, but also more stable for practical real-world deployment.

As shown in Fig.~\ref{fig:qualitative_umetrack_comparision}, UmeTrack, even when using crops derived from ground-truth 3D keypoints, does not reliably recover hand pose under hand--object occlusions (ARCTIC, left).
Even in non-occluded cases, its 2D finger reprojections are less accurate than ours (H2O, middle).
Furthermore, on HOT3D (right), the interacting left hand pose estimate and its reprojection are not faithful to the observed hand.

\begin{table*}[!t]
\centering
\caption{
\textbf{Comparison with UmeTrack on ARCTIC, HOT3D, and H2O (in mm).}
We report camera-space mean joint error (CS-MJE) and Procrustes-aligned mean joint error (PS-MJE).
For UmeTrack, we evaluate two crop settings: 3D keypoints-based perspective crop and 2D bounding-box-based perspective crop.
}
\vspace{-1em}
\small
\begingroup
\setlength{\tabcolsep}{13pt}
\renewcommand{\arraystretch}{1.0}

\begin{tabular}{@{}l
    *{2}{S[table-format=3.1]}
    c
    *{2}{S[table-format=3.1]}
    c
    *{2}{S[table-format=3.1]}
    @{}}
\toprule
Method
& \multicolumn{2}{c}{ARCTIC}
&
& \multicolumn{2}{c}{HOT3D}
&
& \multicolumn{2}{c}{H2O} \\
\cmidrule(lr){2-3}\cmidrule(lr){5-6}\cmidrule(lr){8-9}
& {CS-MJE $\downarrow$} & {PS-MJE $\downarrow$}
&
& {CS-MJE $\downarrow$} & {PS-MJE $\downarrow$}
&
& {CS-MJE $\downarrow$} & {PS-MJE $\downarrow$} \\

\midrule

UmeTrack (3D keypoints crop)
& 55.1 & 14.7
&& \textbf{31.9} & 12.1
&& \textbf{23.8} & 11.1 \\[0.35em]

\midrule

UmeTrack* (2D bounding-box crop)
& 155.4 & 24.5
&& 261.7 & 28.9
&& 124.6 & 21.6 \\

\textsc{EgoForce} (ours)
& \textbf{49.5} & \textbf{8.0}
&& 43.9 & \textbf{6.6}
&& 25.0 & \textbf{5.6} \\

\bottomrule
\end{tabular}
\endgroup
\label{tab:UMETRACK_results}
\end{table*}

\paragraph{HaWoR} 
Since HaWoR depends on SLAM for its global hand pose estimation, it introduces additional computational overhead and makes the pipeline sensitive to failures in camera tracking. 
This is particularly problematic in egocentric hand-interaction scenarios, where the camera is often close to the hands and manipulated objects frequently occupy a large part of the view, reducing the stability of feature matching and pose estimation. 
Such conditions are common in ARCTIC, where close-up views and frequent object interactions make SLAM especially unreliable in our experiments.
As a result, HaWoR performs poorly on ARCTIC, with $319.9$~mm CS-MJE and $16.3$~mm PA-MJE, whereas EgoForce achieves $49.5$~mm CS-MJE and $8.0$~mm PA-MJE. 
On H2O, HaWoR performs better, reaching $72.5$~mm CS-MJE and $6.6$~mm PA-MJE, but EgoForce still outperforms it with $25.0$~mm CS-MJE and $5.6$~mm PA-MJE. 
These results indicate that direct monocular camera-space reconstruction is more robust than SLAM-dependent global pose recovery in close-range egocentric interaction settings.

\label{subsec:intrinsics_robustness}

\noindent
\textbf{Calibration-mismatch robustness.}
In real deployments, intrinsics may be noisy, shifted over time, or only approximately available rather than precisely measured for each device. We therefore evaluate sensitivity to calibration mismatch to test whether performance remains stable under realistic intrinsic errors.
We evaluate calibration-mismatch sensitivity (see Fig.~\ref{fig:intrinsics_robustness}) by interpolating camera intrinsics between the default dataset calibration (0\%) and AnyCalib~\cite{tirado2025anycalib} estimate adapted to HOT3D's camera model (100\%), then extrapolating to 200\%.

To quantify the mismatch, we define camera-geometry error that measures how much the perturbed camera changes viewing rays relative to the reference camera: for a grid of image pixels, we compute ray directions from the reference and perturbed camera models, measure their angular difference, and convert this into a metric displacement at multiple depths to account for near- and far-field hand positions in egocentric capture; the reported value is the mean displacement in millimeters, where larger values indicate stronger geometric inconsistency.
Despite increasing camera-geometry error away from the dataset calibration, hand pose accuracy remains stable over a broad range and is best at 50\%, showing robustness to intrinsic mismatch and shows graceful degradation under extreme deviations (>150\%).
Interestingly, the best result is obtained at an intermediate interpolation between the dataset and AnyCalib intrinsics. However, selecting such an interpolation at deployment is not possible without ground-truth 3D hand poses, motivating future work on combining factory and software-estimated intrinsics for on-the-fly self-calibration without ground-truth.

\label{subsec:depth_scale}

\noindent
\textbf{Arm-based depth--scale stabilization.}
To better understand why arm information improves absolute hand reconstruction, we go beyond the aggregate CS-ACC and RS-MJE gains in Tab.~\ref{tab:arm-cvae-arctic} and analyze its effect on depth--scale ambiguity directly. 
Since monocular egocentric hand estimation is inherently affected by depth--scale ambiguity, we measure hand scale error, defined as the wrist-to-middle-finger MCP distance error, across different hand-to-camera distances.
Arm-based depth--scale stabilization comes from: (1) forearm input to HALO, which provides additional context for improved hand mesh estimation, and (2) the parametric forearm mesh predicted by HALO (FARM shape and pose), which provides the arm 3D joints to RSS and anchors the hand--arm.
Tab.~\ref{tab:depth_scale} shows that arm context helps most in the near field, is neutral in the mid range, and still helps in the far field, supporting its role in mitigating depth--scale ambiguity rather than merely improving temporal smoothness or local mesh quality.

\begin{table}[t]
\caption{\textbf{Arm-based depth--scale stabilization.}
Mean hand scale error on ARCTIC (mm), with changes shown in parentheses. Arm input improves scale estimation, especially in the near field, supporting its role in reducing depth--scale ambiguity.
}
\centering
\small
\setlength{\tabcolsep}{8pt}
\begin{tabular}{lcc}
\toprule
\multirow{2}{*}{\shortstack[l]{Distance of hands\\from camera (mm)}} &
\multicolumn{2}{c}{Mean Hand Scale Error$\downarrow$ (mm)} \\
\cmidrule(lr){2-3}
& Without arm input & With arm input \\
\midrule
200--300 (near-field) & 4.7 & $2.7\,(-2.0\downarrow)$ \\
300--500              & 3.2 & $3.2\,(0.0)$ \\
500--1000 (far-field) & 3.3 & $3.0\,(-0.3\downarrow)$ \\
\bottomrule
\end{tabular}
\label{tab:depth_scale}
\end{table}

\noindent
\textbf{Hand-size metrics.}
To assess whether reconstruction depends on explicit hand-size calibration, we analyze both within-sequence scale stability and generalization across unseen hand sizes. 
Since our method does not use per-user hand-size calibration, we verify that the predicted scale remains stable across frames and is not strongly dependent on subject-specific size tuning.
Even without calibration, frame-wise hand-scale variation remains small: around $4$~mm on HOT3D and $2$~mm on ARCTIC, corresponding to only 4\% and 2\% of the average hand size (HOT3D: $94$~mm and ARCTIC: $90$~mm), respectively. 
Notably, HOT3D and ARCTIC together cover 5 unseen hand sizes. 
Per-sequence calibration further reduces CS-MJE on HOT3D from $43.9$~mm to $42.7$~mm, while yielding no noticeable improvement on ARCTIC.
These results indicate that EgoForce achieves stable scale reconstruction without explicit per-user calibration, is robust to hand-size variation across subjects, and remains stable within a sequence. 
This supports our goal of deployment on smart glasses while keeping additional calibration dependencies minimal.

\subsection{Ablations}

\noindent
\textbf{Camera-Space Lifting.}
Table~\ref{tab:camera_space_lifting} compares four ways for lifting root‐relative predictions into camera space on H2O (pinhole) and HOT3D (fisheye).
The naïve depth-based lifting, similar to \HaMerP’s metric-depth formulation, fails under monocular scale ambiguity, yielding large translation errors (530.5/1851.9$~mm$) and correspondingly high acceleration errors (41.9/180.4$~m/s^2$).
On H2O, our Ray Space Solver (RSS) and DGP from \citet{handdgp2024} achieve the same CS-MJE of $25~mm$ with nearly identical CS-ACC ($7.4~m/s^2$), indicating that both correctly exploit the pinhole projection constraints.
On HOT3D, however, DGP reaches $115.6~mm$ CS-MJE with $25.0~m/s^2$ CS-ACC, whereas RSS gives $45.8~mm$ and $23.5~m/s^2$, respectively.
Applying Kalman filtering on top of RSS (``RSS w/ KF’’) further stabilizes the lifted trajectory, yielding $43.9~mm$  CS-MJE and $14.0~m/s^2$ CS-ACC, the best performance on both datasets.
Please refer to Sup. Fig.~\ref{fig:camera_space_lifting_ablation} for qualitative illustrations of different camera-space lifting techniques, and to the supplementary video for the impact of Kalman filtering on temporal smoothness.

\begin{table}[!h]
  \caption{
    \textbf{Ablation of Camera-Space Lifting Approaches.}
We report CS-MJE (mm) and CS-ACC (m/s$^2$)  on H2O (pinhole) and HOT3D (fisheye).
``Est. Intri'' = Estimated Intrinsics; ``DGP'' = Differential Global Positioning~\cite{handdgp2024}; ``RSS'' = Ray-Space Solver; ``KF'' = Kalman Filtering.
  }
  \centering
  \small
  \begingroup
  \setlength{\tabcolsep}{4pt}
  \begin{tabular}{@{}l
                  *{2}{S[table-format=3.1]}
                  *{2}{S[table-format=3.1]}
                  @{}}
    \toprule
    Method & \multicolumn{2}{c}{H2O} & \multicolumn{2}{c}{HOT3D} \\
    \cmidrule(lr){2-3} \cmidrule(lr){4-5}
           & {CS-MJE $\downarrow$} & {CS-ACC $\downarrow$}
           & {CS-MJE $\downarrow$} & {CS-ACC $\downarrow$} \\
    \midrule
    HALO + Depth        & 530.5 & 41.9 & 1851.9 & 180.4 \\
    HALO + DGP          &  25.0 &  7.4 &  115.6 &  25.0 \\
    HALO + RSS (w/o KF)   & 25.0 & 7.4 & 45.8 & 23.5 \\
    HALO + RSS (w KF) (Ours)   & \best{25.0} & \best{5.5} & \best{43.9} & \best{14.0} \\
    \specialrule{.7pt}{4pt}{2pt}
  \end{tabular}
  \endgroup
  \label{tab:camera_space_lifting}
\end{table}

\noindent
\textbf{Isolating CIT from undistortion.}
In Tab.~\ref{tab:ablation_CIT_2}, we keep undistortion fixed and toggle only CIT. 
Across all radial regions, CIT consistently reduces hand CS-MJE. Importantly, these gains persist even after undistortion, confirming that CIT provides benefits beyond preprocessing. The improvements remain consistent across the full image, including peripheral regions where fisheye distortion is strongest, indicating that CIT improves robustness throughout the image, consistent with Tab.~\ref{tab:ablation_CIT}.

\begin{table}[t]
\caption{\textbf{Ablation of CIT.}
Hand CS-MJE on HOT3D (mm), with the change shown in parentheses. Keeping undistortion fixed, CIT consistently improves performance.}
\label{tab:ablation_CIT_2}
\vspace{-0.5em}
\centering
\small
\setlength{\tabcolsep}{5pt}
\begin{tabular}{lcl}
\toprule
\shortstack[l]{Hand location\\by radial region (\%)} 
& \shortstack[c]{With undistortion\\w/o CIT $\rightarrow$ w/ CIT}
& \shortstack[l]{Without undistortion\\w/o CIT $\rightarrow$ w/ CIT} \\
\midrule
0--25     & $38.7 \rightarrow 36.2\,(-2.5\downarrow)$  & \hspace{1pt} $67.4~ \rightarrow 42.8\,(-24.6\downarrow)$ \\
25--50    & $42.5 \rightarrow 38.8\,(-3.8\downarrow)$  & \hspace{1pt} $95.7~ \rightarrow 56.8\,(-38.9\downarrow)$ \\
50--75    & $43.5 \rightarrow 40.4\,(-3.1\downarrow)$  & \hspace{1pt} $141.5 \rightarrow 82.1\,(-59.4\downarrow)$ \\
$\geq 75$ & $58.5 \rightarrow 54.8\,(-3.7\downarrow)$  & \hspace{1pt} $152.8 \rightarrow 99.3\,(-53.5\downarrow)$ \\
\bottomrule
\end{tabular}
\end{table}

\section{Implementation Details}
\label{supp:sec:extended_implementation}

\noindent\textbf{Arm--Hand Crop Encoder.}
Given the hand and arm crops $\mathbf I_H\in\mathbb{R}^{224\times224\times3}$ and $\mathbf I_A\in\mathbb{R}^{112\times112\times3}$, patchification with patch size $p=16$ produces
$N_H=(224/p)^2=14^2=196$ hand tokens and $N_A=(112/p)^2=7^2=49$ arm tokens, for a total of $N=N_H+N_A=245$ tokens.
We use a ViT-H/16 backbone (pretrained ViTPose-H weights) with token and feature dimension $d=c=1280$.
This yields hand tokens $\mathbf T_H\in\mathbb{R}^{196\times1280}$ and arm tokens $\mathbf T_A\in\mathbb{R}^{49\times1280}$ and concatenating them produces the encoded visual tokens $\mathbf X\in\mathbb{R}^{245\times1280}$.
The Crop Intrinsics Tokens have dimension $k=128$ (Sec.~\ref{sec:CIT}) and are fused per patch by concatenation, a linear projection $\mathbb{R}^{d+k}\!\rightarrow\!\mathbb{R}^{d}$, and a residual addition (see Fig.~\ref{fig:CaDBlock}).

\noindent\textbf{Contextual Decoding of Hand--Arm Interactions.}
We instantiate four hand queries and three arm queries,
$\mathbf Q_H\in\mathbb{R}^{4\times c}$ (2D joints, global pose, hand shape, hand pose) and
$\mathbf Q_A\in\mathbb{R}^{3\times c}$ (2D joints, arm shape, arm pose),
and stack them to form the target sequence
$\mathbf Q_0=[\mathbf Q_H;\mathbf Q_A]\in\mathbb{R}^{7\times c}$.
A transformer decoder with $L_{\mathrm{dec}}=2$ layers and $h_{\mathrm{dec}}=8$ attention heads attends to the encoded patch tokens $\mathbf X\in\mathbb{R}^{N\times c}$ and outputs
$\mathbf Q_L\in\mathbb{R}^{7\times r}$, where $r=1280$.
We split $\mathbf Q_L$ into hand and arm features,
$\mathbf f_{\mathrm{hand}}\in\mathbb{R}^{4\times1280}$ and $\mathbf f_{\mathrm{arm}}\in\mathbb{R}^{3\times1280}$.
The decoder self-attention enables information exchange between hand and arm queries, while cross-attention grounds each query in the visual evidence provided by $\mathbf X$.

\noindent\textbf{Plausible Arm Completion.}
When the forearm is not visible, we replace the missing arm query features using a hand-conditioned variational prior.
Specifically, we predict $(\mu,\log\sigma^2)\in\mathbb{R}^{128}$ from the available hand features using two linear heads, sample a latent arm code $\mathbf z_{\mathrm{arm}}\in\mathbb{R}^{128}$ via the reparameterization trick, and project it to the feature width $D=1280$.
We then decode this embedding using three residual MLP blocks (LayerNorm + ReLU).
A final linear layer outputs an arm-query embedding $\mathbb{R}^{3\times 1280}$ and is used to inpaint the missing arm query features.

\noindent\textbf{2D Joint Decoder.}
We decode $56\times56$ heatmaps for $J_H=21$ hand joints and $J_A=3$ forearm joints, and obtain 2D coordinates via soft-argmax using $\mathrm{softmax}(\tau\,\mathbf H)$ with learnable temperature $\tau$ (initialized to $1$).
Per-joint confidence weights $w_j\in(0,1)$ are predicted by an MLP on features bilinearly sampled at the predicted joint locations.

\noindent\textbf{Ray Space Solver.}
We estimate the camera-space translation $\mathbf t\in\mathbb{R}^3$ from $24$ 2D--3D correspondences (21 hand + 3 forearm joints).
Let $\tilde{\mathbf u}_i=(\tilde u_i,\tilde v_i)$ be the predicted 2D joint in crop coordinates, where the network input crop has size $(W_{\mathrm{in}},H_{\mathrm{in}})$.
We map crop coordinates back to full-image pixels by
$$
u_i = x_0 + s_x\,\frac{\tilde u_i}{W_{\mathrm{in}}},\qquad
v_i = y_0 + s_y\,\frac{\tilde v_i}{H_{\mathrm{in}}},
$$
where $(x_0,y_0)$ denotes the top-left corner of the crop's bounding box in the full image, and $(s_x,s_y)$ denotes its size in full-image pixels (width and height), \ie, $s_x=x_1-x_0$ and $s_y=y_1-y_0$ for the bottom-right corner $(x_1,y_1)$.
These full-image joint coordinates are normalized by calibrated intrinsics and unprojected through the native camera model to form unit bearing rays, after which we solve for $\mathbf t$ via a weighted point-to-ray least-squares system in closed form (see Sec.~\ref{sec:ray_depth_solve}).

\noindent\textbf{Kalman Filter.}
We temporally smooth the estimated camera-space translation $\mathbf t$ using a constant-velocity Kalman filter at $\text{freq}=30$~Hz, with process-noise variances $q_{\mathrm{pos}}=0.001$ (position) and $q_{\mathrm{vel}}=10^{-5}$ (velocity), and measurement-noise variance $r_{\mathrm{meas}}=0.001$.

\section{Limitations}
\label{supp:sec:extended_limitations}
Beyond the limitations discussed in the paper, we note several additional considerations.

\smallskip 

\noindent (1) Although the forearm prior mitigates monocular depth--scale ambiguity, recovering a precisely metrically scaled MANO hand--arm configuration from a single monocular device remains underconstrained without user-specific scale cues (e.g., hand size or limb length).

\smallskip

\noindent (2) The arm modeling primarily serves as a geometric prior to stabilize and improve hand pose estimation. 
While the predicted arm meshes are often plausible, they are not yet optimized for tasks requiring precise hand--forearm localization.

\smallskip 

\noindent (3) More generally, inferring the individual limb geometry remains difficult under severe occlusion, limited field of view, and fast motion. 
A more holistic model that jointly reasons about both the limbs, or even the upper body, could provide a stronger kinematic context and further stabilize hand and arm estimates.

\end{document}